\documentclass{article}

\PassOptionsToPackage{table}{xcolor}
\usepackage{iclr2027_conference,times}

\usepackage{microtype}
\usepackage{graphicx}
\usepackage{wrapfig}
\usepackage{flafter}
\usepackage{subcaption}
\usepackage{booktabs}
\usepackage{multirow}
\usepackage{array}
\usepackage{tabularx}
\usepackage{pdflscape}
\usepackage{algorithm}
\usepackage{algorithmic}
\usepackage{hyperref}
\hypersetup{hidelinks}
\usepackage{url}
\usepackage{amsmath,amssymb,mathtools,amsthm}
\usepackage[capitalize,noabbrev]{cleveref}
\usepackage{xspace}
\usepackage{enumitem}
\usepackage{siunitx}
\usepackage{needspace}
\usepackage{placeins}
\usepackage{xcolor,soul}
\usepackage{wrapfig}

\newcommand{\relu}[1]{\left[#1\right]_{+}}
\newcolumntype{L}[1]{>{\raggedright\arraybackslash}p{#1}}
\newcolumntype{C}[1]{>{\centering\arraybackslash}p{#1}}
\newcolumntype{Y}{>{\centering\arraybackslash}X}

\definecolor{tablehead}{gray}{0.88}
\definecolor{tablegroup}{gray}{0.93}
\definecolor{tablerow}{gray}{0.97}

\theoremstyle{plain}

\theoremstyle{definition}

\theoremstyle{remark}

\title{Not Too Hard, Not Too Easy: Learning from Intermediate States for LLM Structured Reasoning}

\author{Hongbo Chen$^{1}$ \quad Guohua Lu$^{1}$ \quad Ting Dang$^{2}$ \quad Hong Jia$^{1}$ \\
\normalfont $^{1}$University of Auckland, Auckland, New Zealand \\
\normalfont $^{2}$University of Melbourne, Melbourne, Australia \\
\normalfont\texttt{hche322@aucklanduni.ac.nz}, \texttt{glu727@aucklanduni.ac.nz} \\
\normalfont\texttt{ting.dang@unimelb.edu.au}, \texttt{hong.jia@auckland.ac.nz}}

\iclrfinalcopy

\begin{document}

\maketitle
\fancyhead[L]{Preprint}

\begin{abstract}

A common principle of effective learning is to practice material that is neither already mastered nor too difficult to permit progress. We ask how to apply this principle to structured reasoning tasks such as Sudoku and maze solving. In these tasks, a model can repeatedly revise an incomplete or incorrect candidate solution until it satisfies the problem’s constraints. The intermediate candidate solutions along this trajectory provide natural training examples: some are already solved, some cannot yet be repaired by the model, and others lie at its current frontier of achievable progress. We therefore investigate whether pretrained language models can learn to revise such states and whether training on states at this frontier improves reasoning more broadly. To achieve this, we couple a pretrained language-model backbone with a recurrent updater that repeatedly revises an explicit solution state, using the same parameters at every update step. We further introduce \textbf{F}rontier-\textbf{O}riented \textbf{C}uration \textbf{U}sing \textbf{S}elf-trajectories (\textbf{FOCUS}), which selects training states from trajectories generated by the current model. FOCUS measures how much the model improves each state within a fixed number of recurrent updates and prioritizes states from which it can make substantial progress. With Qwen3-1.7B, FOCUS achieves 64.4\% exact solve accuracy on Sudoku-Extreme and 91.1\% on Maze-Hard, with similar gains observed across five Qwen and Llama backbones spanning 1.7B to 8B parameters. We further observe zero-shot transfer in the adapted LLM to mathematical reasoning and code execution, even when the recurrent updater is disabled and no downstream fine-tuning is performed. 


\end{abstract}

\section{Introduction}

Learning depends not only on the amount of practice, but also on the difficulty of the material being practiced. Under specific learning assumptions, examples of intermediate difficulty can support faster learning than examples that are too easy or too hard
\citep{wilson2019eightyfive}. Human learners also use their recent progress to choose what to practice next \citep{ten2021learningprogress}. This leads to a basic, yet less investigated, question for reasoning models: \emph{which training examples are most useful at the model's current level of ability?}


We study this question on Sudoku-Extreme and Maze-Hard, two structured reasoning tasks on which prior evaluations report low exact-solve accuracy for language models \citep{wang2025hrm,jolicoeur2025trm}. Both tasks have explicit solutions that can be checked at every step. A Sudoku grid must satisfy its row, column, and box constraints, while a maze path must connect the start and goal through traversable cells. A model can therefore revise a candidate solution repeatedly, and we can measure whether each revision moves it closer to the correct answer. Because each intermediate solution can be checked, these tasks let us ask which intermediate solutions are most useful for training.
To study this question, we couple a pretrained language model with a recurrent updater that produces a sequence of revisable intermediate states. The language model encodes the problem, and the updater repeatedly revises an explicit answer using the same parameters at every step. Each intermediate state contains the current answer and the updater's memory. Standard training begins from the initial state. We can also save an intermediate state and resume training from it. The question then becomes not only which problems to use, but also \emph{where in each solution sequence to resume training}.

Prior approaches use several sources of training states. Training can begin from the initial state used at inference, an artificially corrupted target, or a state visited by the current model. However, these options do not explain which state within a model-generated sequence is most useful for additional training.


We address this problem with Frontier-Oriented Curation Using Self-trajectories (FOCUS). During training, FOCUS records the states visited by the current solver. From each candidate state, it measures the proportional reduction in task error over a fixed number of recurrent steps. It then selects the state whose reduction is closest to a chosen target. Here, difficulty is defined by short-term repair progress: states that improve much more than the target are relatively easy, states that improve much less are relatively hard, and states near the target form the \emph{repair frontier}. Because FOCUS measures progress using the current solver, the frontier changes as the model improves. Figure~\ref{fig:focus_overview} summarizes the procedure.

After selecting a state, FOCUS resumes training from both its saved answer and its saved memory. The training loss penalizes the model when its error reduction falls short of the target. For compatible updates that start from the original state, a frozen TRM teacher also provides soft prediction targets at selected steps \citep{hinton2015distillingknowledgeneuralnetwork}. At inference, the language model encodes the problem once, the updater runs for a fixed number of steps, and the model decodes only the final state. State selection, teacher predictions, and training-time error scores are not used at inference.


With Qwen3-1.7B, FOCUS achieves 64.4\% exact accuracy on Sudoku-Extreme and 91.1\% on Maze-Hard, compared with 7.2\% and 86.0\% for Recurrent final-only. FOCUS outperforms this baseline on both tasks across all five Qwen and Llama backbones ranging from 1.7B to 8B parameters (Table~\ref{tab:source_task_compact}). Appendix~\ref{app:backbone_portability} describes the evaluation protocol.

We also test whether language-model parameters learned during Sudoku training also improve reasoning performance on mathematical and code-reasoning tasks. To do so, we disable the recurrent updater and evaluate the adapted language model on these tasks \textit{without additional training}. Compared with the original Qwen3-1.7B model, the Sudoku-trained adapter improves zero-shot accuracy from 72.36\% to 72.96\% on MATH-Hard and from 50.75\% to 55.38\% on CruxEval-O.




Our contributions are threefold:
\begin{enumerate}[leftmargin=*,label=\textbullet]
    \item We show how a pretrained language model can guide a recurrent updater that repeatedly revises an explicit answer. We compare direct prediction, one update, and repeated updates across two structured tasks and five language-model backbones.

    \item We introduce FOCUS, which selects intermediate states by measuring how much the current model improves them over a fixed number of steps. Training resumes from the selected answer and memory and penalizes progress below a chosen target.

    \item We provide a controlled evaluation of whether FOCUS changes the language model beyond its training tasks. We show that language-model adapters trained with FOCUS can improve zero-shot performance on different mathematical and code-reasoning benchmarks.
\end{enumerate}

\section{Related Work}


\paragraph{Iterative reasoning.}
Prior work develops specialized recurrent architectures for structured reasoning. The Hierarchical Reasoning Model (HRM) uses recurrent modules operating at different timescales, while the Tiny Recursive Model (TRM) uses a smaller recursive network to revise solutions repeatedly~\citep{wang2025hrm,jolicoeur2025trm}. The Denoising Recursion Model (DRM) trains a recursive network to recover correct solutions from deliberately corrupted ones~\citep{cameron2026drm}. These models demonstrate the value of repeated solution updates, but they focus on model architecture and repair training rather than selecting states from the current updater’s trajectory based on measured progress. Unlike these specialized recurrent architectures, FOCUS couples a task-specific recurrent updater with a pretrained language model and is compatible with multiple language-model backbones.


\paragraph{Training-state selection.}
Prior methods use loss~\citep{shrivastava2016ohem}, predefined difficulty~\citep{parashar2026e2h}, success rates~\citep{florensa2017reverse,florensa2018goalgan}, or value-estimation errors~\citep{jiang2021plr} to decide what the model should practice. These signals select examples, starting states, goals, or environments, but they do not directly measure how repeated updates change a particular intermediate solution. As such, initial error is insufficient for this purpose because two states with similar errors may respond differently to the same updates. FOCUS instead runs the current updater for a fixed number of steps from each candidate state and measures the proportional reduction in task error. An observed reduction larger than the target exceeds the desired progress, whereas a smaller reduction falls short. FOCUS selects the state whose reduction is closest to this target, called the repair frontier, rather than selecting by initial error or by the largest observed improvement.

\section{Method}
\label{sec:method}

\paragraph{Overview.}

\begin{figure}[!htbp]
  \centering
  \includegraphics[width=\textwidth]{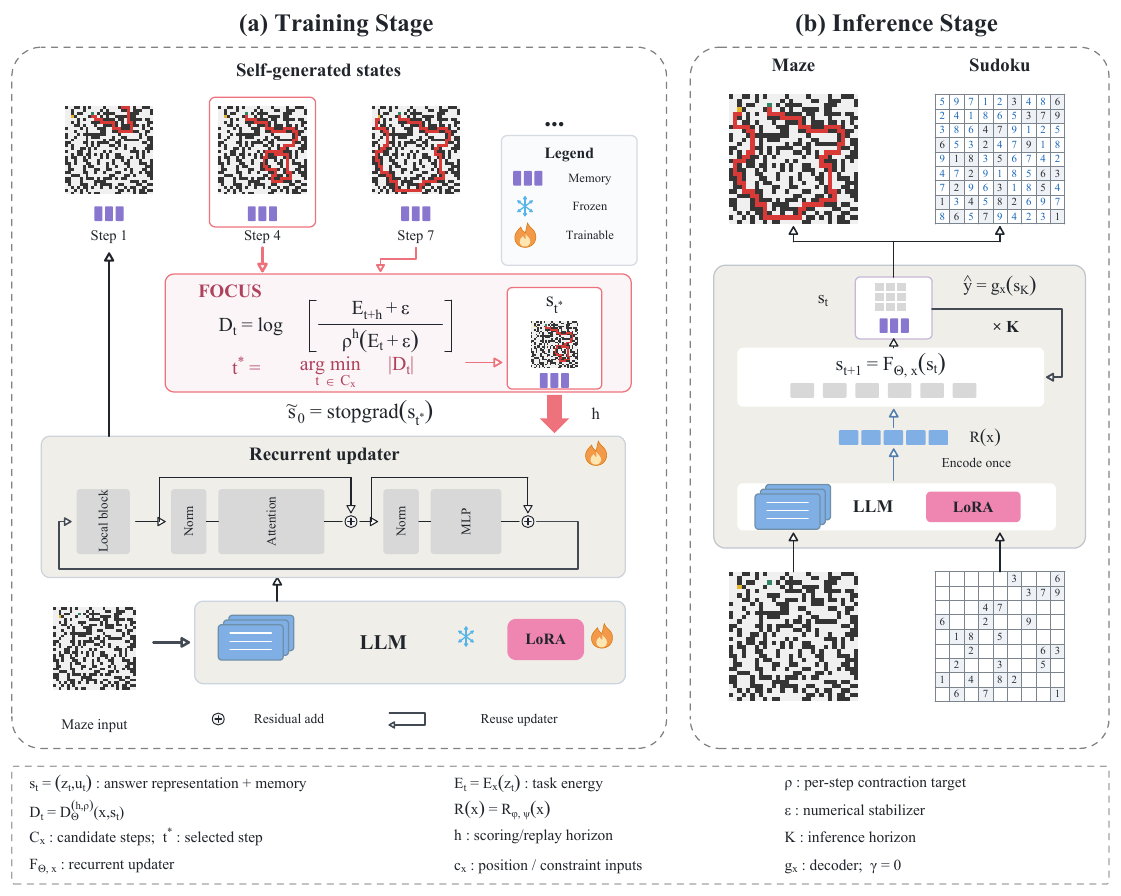}
\caption{Overview of FOCUS.
(a) Training selects intermediate states near the repair frontier and resumes recurrent updates from the selected answer and memory.
(b) At inference, the adapted LLM encodes the input once to condition $K$ recurrent updates, followed by a single final decode.}
  \label{fig:focus_overview}
\end{figure}

Figure~\ref{fig:focus_overview} summarizes FOCUS. A pretrained language model encodes each problem, and a recurrent updater uses this representation to revise an explicit answer over successive steps. Training from the task-defined initial state produces a trajectory of intermediate states. FOCUS selects one state on this trajectory, restores both its answer and recurrent memory, and trains the updater to continue from it. The selection criterion measures how much the current updater reduces task error over a fixed number of steps and chooses the state whose reduction is closest to a prescribed target. Training from the original initialization remains part of the procedure. At inference, the updater starts from this initial state and applies a fixed number of updates.

\subsection{Language-Model-Conditioned Recurrent Updates}
\label{sec:recurrent_updater}

For an input \(x\), the model maintains a state \(s_t=(z_t,u_t)\). The component \(z_t\) is the current answer, and \(u_t\) is the memory carried between updates. Training uses the correct answer \(y^\star\). A pretrained language model encodes \(x\), and a task-specific projection maps the encoding to
\begin{equation}
R_{\phi,\psi}(x)
=
G_\psi\!\left(
\operatorname{LM}_{W_0+\Delta W_\phi}(x)
\right).
\label{eq:task_representation}
\end{equation}
The pretrained weights \(W_0\) remain fixed. The low-rank adaptation parameters \(\phi\) and the projection parameters \(\psi\) are trained \citep{hu2022lora}. The representation \(R_{\phi,\psi}(x)\) supplies the problem context, while \(s_t\) records the answer being revised. It is computed once and reused at every update.

The trainable parameters are \(\Theta=(\theta,\phi,\psi)\). A recurrent updater, separate from the language model, computes
\begin{equation}
s_{t+1}
=
F_{\Theta,x}(s_t)
:=
F_\theta\!\left(
s_t;R_{\phi,\psi}(x),c_x
\right),
\label{eq:recurrent_update}
\end{equation}
where \(c_x\) contains deterministic positional and task-constraint inputs. The updater parameters \(\theta\) are shared across steps. Gradients from these updates reach \(\theta\), \(\phi\), and \(\psi\).

At inference, the language model encodes \(x\) once. The updater starts from \(s_0^{\mathrm{clean}}(x)\), which is independent of \(y^\star\), and applies \(K\) updates. The prediction is \(\hat y=g_x(s_K)\). Appendix~\ref{app:implementation} specifies the task-specific states and initialization, and Appendix~\ref{app:evaluation} specifies decoding.

\subsection{Frontier-Based State Selection}
\label{sec:finite_horizon_progress}
\label{sec:focus_selection}

\paragraph{Measuring repair behavior.}
A state's current error indicates how much work remains. Its continuation
shows how the updater handles that work. States with similar error can
improve at different rates under the same number of updates. FOCUS
therefore measures energy reduction relative to starting energy over a
fixed horizon.

\paragraph{Candidate states.}
With gradient tracking disabled, the current updater generates a trajectory
\(\tau(x)=(s_0^{\mathrm{clean}},s_1,\ldots,s_{K_p})\). Here \(K_p\) is the
collection horizon. Scoring a state \(s_t\) requires its state after \(h\)
updates. Both states appear in the collected trajectory when
\(0\leq t\leq K_p-h\). The candidate set \(\mathcal C_x\) contains the
indices in this range that also satisfy the task-specific eligibility
rules in Appendix~\ref{app:focus_candidates}. That appendix also specifies
tie-breaking. The horizon \(h\) is used for both scoring and replay, while
\(K\) is the inference horizon. Parameters remain fixed during collection
and scoring.

\paragraph{Contraction defect.}
Let \(E_x(z)\geq0\) be a task energy combining supervised error with
task-specific penalties, as defined in Appendix~\ref{app:energy}. Lower
energy indicates improvement under this surrogate. It may use \(y^\star\),
so self-generated states do not imply self-generated supervision. For
\(s=(z,u)\), write \(z^{(h)}=\pi_zF_{\Theta,x}^{h}(s)\), where \(\pi_z\)
extracts the answer component. We define
\begin{equation}
D_{\Theta}^{(h,\rho)}(x,s)
=
\log\!\left(
\frac{E_x(z^{(h)})+\epsilon}
{\rho^h\bigl(E_x(z)+\epsilon\bigr)}
\right),
\qquad
0<\rho<1,\quad \epsilon>0.
\label{eq:defect}
\end{equation}
The reference factor \(\rho^h\) specifies a target reduction over \(h\)
steps, not a rate assumed to hold for the learned updater. A nonpositive
defect meets this target for the smoothed energy \(E_x+\epsilon\); a positive
defect falls short, even when energy decreases. The logarithm expresses
multiplicative deviations as signed additive quantities, which accumulate
across consecutive windows (\textit{cf}. Appendix~\ref{app:proof}).

\paragraph{Selecting the repair frontier.}
Defects near zero identify continuations near the target reduction. Small
positive values indicate a small shortfall, while small negative values
indicate that the target has just been met. FOCUS selects from either side
of this boundary. For a nonempty candidate set, it chooses
\begin{equation}
t^\star
=
\operatorname*{arg\,min}_{t\in\mathcal C_x}
\left|D_{\Theta}^{(h,\rho)}(x,s_t)-\gamma\right|.
\label{eq:focus_selection}
\end{equation}
We use \(\gamma=0\) and call states near this level the \emph{repair
frontier}. The closest eligible state is selected even if its defect is
far from zero. As the updater changes, different states can become
closest to the target.

\subsection{Replaying Complete Recurrent States}
\label{sec:training_objective}

\paragraph{Restoring the selected state.}
A continuation depends on both the answer \(z_t\) and memory \(u_t\).
FOCUS restores the complete selected state and trains an \(h\)-step
continuation
\begin{equation}
\widetilde s_0
=
\operatorname{stopgrad}(s_{t^\star}),
\qquad
\widetilde s_{j+1}
=
F_{\Theta,x}(\widetilde s_j),
\quad j=0,\ldots,h-1.
\label{eq:full_state_replay}
\end{equation}
The saved state is detached, so backpropagation covers the replay
continuation. The conditioning representation is recomputed with gradient
tracking. Replay can therefore update \(\theta\), \(\phi\), and \(\psi\).

\paragraph{Task supervision and local reduction.}
Task supervision trains the replayed continuation toward the correct
answer. An additional term penalizes shortfalls relative to the selected
state's contraction target. Let \(D^{\mathrm{rep}}\) denote
\eqref{eq:defect} recomputed on the gradient-tracked replay. The replay
objective is
\begin{equation}
\mathcal L_{\mathrm{aux}}^{\mathrm{FOCUS}}
=
\mathcal L_{\mathrm{task}}
(\widetilde s_{1:h};x,y^\star)
+
\beta_{\mathrm{def}}[D^{\mathrm{rep}}-\gamma]_+^2,
\label{eq:cfc_loss}
\end{equation}
where \([a]_+=\max(a,0)\) and \(\beta_{\mathrm{def}}\geq 0\). The
one-sided penalty applies when replay falls short of the target. Task
supervision applies throughout the replay continuation. Gradients through
the defect term follow the replayed updates. The task loss and energy may
have different components and temporal reductions, as specified in
Appendix~\ref{app:implementation}.

\paragraph{Training from the original initialization.}
Replay trains continuations from visited states. We also train rollouts
from the task-defined initial state used at evaluation. On compatible
teacher-supervised initial-state updates, temperature-scaled KL supervision
matches the student's soft predictions to those of a frozen TRM teacher
at aligned recurrent steps. Teacher supervision applies to initial-state
updates.

Sudoku includes a base loss in every mini-batch and adds replay loss when
scheduled. Maze-Hard alternates between initial-state and replay updates.
When \(\mathcal C_x\) is empty, Sudoku omits the replay term and Maze-Hard
uses an initial-state update. Branch weights, teacher alignment, the Fixed
Mix baseline, and the complete algorithm are specified in
Appendix~\ref{app:implementation}. At inference, the model follows the
fixed rollout defined in Section~\ref{sec:recurrent_updater}.

\Needspace{5\baselineskip}
\section{Experiments}
\label{sec:experiments}

We first evaluate how FOCUS can solve Sudoku-Extreme and Maze-Hard problems. Then we examine zero-shot transfer performance on code and math tasks, followed by a discussion of training-state curation, ablations, and qualitative analysis of recurrent trajectories.

\subsection{Experimental Setup}
\label{sec:experimental_setup}

\paragraph{Tasks and data.}
Sudoku-Extreme uses an $81\times9$ answer-logit grid with given digits clamped after each update~\citep{jolicoeur2025trm}. Training uses symmetry augmentations of 707 source puzzles, supplemented with synthetic puzzles. Both training corpora are puzzle-disjoint from the 1,000 validation and 1,000 held-out test problems. Maze-Hard uses $900/100/1{,}000$ train/validation/test mazes. Appendix~\ref{app:evaluation} provides the data construction details and output formats.


\paragraph{Models and training configurations.}
We evaluate Qwen3-1.7B, Qwen3-4B, Qwen3-8B, Llama-3.2-3B-Instruct, and Llama-3.1-8B-Instruct \citep{yang2025qwen3technicalreport, meta2024llama32, grattafiori2024llama3herdmodels}. Base model uses the pretrained language model without task-specific training. Direct LoRA SFT trains the model to generate answers directly. One-step State-SFT supervises a single explicit-state update. Recurrent final-only supervises the final answer after repeated shared updates. FOCUS follows the training procedure in Section~\ref{sec:method}. All experiments used a fixed seed of 42. The curation study uses Qwen3-1.7B. Teacher-supported configurations apply TRM distillation to compatible initial-state updates. Appendix~\ref{app:baselines} gives more detailed training configurations.

\paragraph{Metrics.}

\begin{table}[t]
\centering
\caption{\textbf{Source-task accuracy across language-model backbones.} Exact solve accuracy (\%) on Sudoku-Extreme and Maze-Hard.}
\label{tab:source_task_compact}
\footnotesize
\setlength{\tabcolsep}{1.5pt}
\renewcommand{\arraystretch}{1.15}
\begin{tabularx}{\linewidth}{@{}L{3.35cm}*{4}{Y}@{\hspace{5pt}}*{4}{Y}@{}}
\toprule
\textbf{Backbone}
& \multicolumn{4}{c}{\textbf{Sudoku-Extreme}}
& \multicolumn{4}{c}{\textbf{Maze-Hard}} \\
\cmidrule(lr){2-5}\cmidrule(lr){6-9}
& \scriptsize Direct
& \scriptsize One-step
& \scriptsize Recurrent
& \cellcolor{gray!6}\scriptsize\textbf{FOCUS}
& \scriptsize Direct
& \scriptsize One-step
& \scriptsize Recurrent
& \cellcolor{gray!6}\scriptsize\textbf{FOCUS} \\
\midrule
Qwen3-1.7B & 0.5 & 0.5 & 7.2 & \cellcolor{gray!6}\textbf{64.4} & 0.0 & 0.0 & 86.0 & \cellcolor{gray!6}\textbf{91.1} \\
Qwen3-4B & 0.5 & 0.3 & 16.2 & \cellcolor{gray!6}\textbf{66.0} & 0.1 & 0.0 & 86.9 & \cellcolor{gray!6}\textbf{91.5} \\
Qwen3-8B & 0.6 & 0.7 & 14.8 & \cellcolor{gray!6}\textbf{64.7} & 0.0 & 0.0 & 87.2 & \cellcolor{gray!6}\textbf{88.3} \\
\addlinespace[2pt]
Llama-3.2-3B-Instruct & 0.0 & 0.4 & 12.3 & \cellcolor{gray!6}\textbf{61.9} & 0.0 & 0.0 & 86.8 & \cellcolor{gray!6}\textbf{90.1} \\
Llama-3.1-8B-Instruct & 0.3 & 0.8 & 8.0 & \cellcolor{gray!6}\textbf{65.8} & 0.0 & 0.0 & 79.8 & \cellcolor{gray!6}\textbf{87.7} \\
\bottomrule
\end{tabularx}
\par\smallskip
\begin{minipage}{\linewidth}
\footnotesize
\raggedright
Direct: Direct LoRA SFT; One-step: One-step State-SFT; Recurrent: Recurrent final-only.
All unadapted base models score 0.0\% on both tasks.
\end{minipage}
\end{table}
In Sudoku and Maze, a problem counts as solved only when the entire predicted answer matches the correct answer. We select models using validation accuracy. For the depth analysis, we also select the number of recurrent updates on the validation set. We report final accuracy on a separate, held-out test set. At test time, the model starts from the task-defined initial state, makes a fixed number of updates, and produces one answer. Each problem receives one attempt without search, teacher assistance, or feedback from a checker. Scoring and selecting intermediate states, then training from them, take place only during training.

\label{sec:source_task_capability}
\subsection{Results}

\begin{wraptable}{R}{0.45\linewidth}
\vspace{-6pt}
\centering
\caption{\textbf{Prompt-only reasoning models on source tasks.}
Exact solve accuracy (\%).}
\label{tab:source_task_prompt_only}
\footnotesize
\setlength{\tabcolsep}{4pt}
\begin{tabular}{@{}lcc@{}}
\toprule
\textbf{Model} & \shortstack{\textbf{Sudoku-}\\\textbf{Extreme}} & \shortstack{\textbf{Maze-}\\\textbf{Hard}} \\
\midrule
DeepSeek-R1    & 0.0 & 0.0 \\
Claude 3.7 8K  & 0.0 & 0.0 \\
o3-mini (high) & 0.0 & 0.0 \\
\bottomrule
\end{tabular}
\vspace{-10pt}
\end{wraptable}

Table~\ref{tab:source_task_compact} reports results on Sudoku-Extreme and Maze-Hard. With Qwen3-1.7B, FOCUS achieves $64.4\%$ on Sudoku-Extreme and $91.1\%$ on Maze-Hard, compared with $7.2\%$ and $86\%$ for Recurrent final-only. Direct LoRA SFT and One-step State-SFT each achieve $0.5\%$ on Sudoku-Extreme and $0$ accuracy on Maze-Hard. Repeated updates also improve solving performance over direct and one-step prediction on the other backbones. Table~\ref{tab:source_task_prompt_only} shows that all three prompt-only reasoning models achieve $0$ accuracy on both tasks. Appendix~\ref{app:backbone_portability} provides evaluation details.

In sum, training recurrence using only final-answer supervision solves most Maze instances but performs substantially worse on Sudoku. In comparison, FOCUS achieves higher accuracy on both tasks. Section~\ref{sec:matched_results} further discusses curation details to explain this difference.

\subsection{Zero-Shot Transfer}
\label{sec:transfer}
\label{sec:downstream_transfer}



\begin{wraptable}{r}{0.53\linewidth}
\centering
\vspace{-2.5em}
\captionsetup{font=small,justification=raggedright,singlelinecheck=false}
\caption{Zero-shot transfer to AIME25 and AIME26 after
structured-repair training.}
\label{tab:downstream_aime}

\footnotesize
\setlength{\tabcolsep}{4pt}
\renewcommand{\arraystretch}{1}

\begin{tabularx}{\linewidth}{@{}>{\raggedright\arraybackslash}Xcc@{}}
\toprule
\rowcolor{tablehead}
\shortstack[l]{\textbf{Tasks}}
  & \textbf{AIME25} & \textbf{AIME26} \\
\midrule
\rowcolor{tablegroup}
\multicolumn{3}{@{}l}{\textbf{Qwen3-1.7B}} \\
Baseline       & 23.33          & 16.67 \\
FOCUS (Sudoku) & 16.67          & \textbf{23.33} \\
FOCUS (Maze)   & \textbf{26.67} & 13.33 \\
\addlinespace[2pt]
\rowcolor{tablegroup}
\multicolumn{3}{@{}l}{\textbf{Qwen3-4B}} \\
Baseline       & 20.00          & 30.00 \\
FOCUS (Sudoku) & 26.67          & \textbf{36.67} \\
FOCUS (Maze)   & \textbf{30.00} & 33.33 \\
\addlinespace[2pt]
\rowcolor{tablegroup}
\multicolumn{3}{@{}l}{\textbf{Qwen3-8B}} \\
Baseline       & 26.67          & 30.00 \\
FOCUS (Sudoku) & 20.00          & 30.00 \\
FOCUS (Maze)   & \textbf{33.33} & \textbf{33.33} \\
\bottomrule
\end{tabularx}

\end{wraptable}

We disable the recurrent module and evaluate the language-model adapters trained on the source tasks using zero-shot CoT prompts and one greedy completion per item. Evaluation uses neither worked demonstrations nor downstream parameter updates \citep{NEURIPS2022_8bb0d291}. Table~\ref{tab:downstream_all_reasoning} reports results across all backbones. Appendix~\ref{app:downstream} describes model selection and downstream evaluation settings.


For Qwen3-1.7B, FOCUS (Sudoku) raises MATH-Hard accuracy from $72.36\%$ to $72.96\%$, CruxEval-O from $50.75\%$ to $55.38\%$, and CruxEval-I from $24\%$ to $26.75\%$ \citep{hendrycks2021math, gu2024cruxeval}.

MATH-Hard contains the 1,324 Level-5 problems from the MATH test set \citep{hendrycks2021math, lighteval2024mathhard}. AIME25 \citep{testtimecompute2025aime} and AIME26 \citep{dekoninck2026matharena} each contain 30 problems. Table~\ref{tab:downstream_aime} reports the AIME results, which vary across backbones and source tasks.

Appendix~\ref{app:downstream_qualitative} compares baseline and FOCUS responses on mathematical reasoning and code-execution tasks.

\begin{table}[!htbp]
\centering
\caption{Zero-shot transfer after structured-repair training. Accuracy (\%) with one greedy completion per item. The recurrent module is disabled and no downstream parameters are updated.}
\label{tab:downstream_all_reasoning}
\footnotesize
\setlength{\tabcolsep}{2pt}
\renewcommand{\arraystretch}{1.12}
\begin{tabularx}{\textwidth}{@{}L{3.0cm}*{5}{Y}@{}}
\toprule
& \multicolumn{2}{c}{\textbf{Mathematics}}
& \multicolumn{2}{c}{\textbf{Code execution}}
& \\
\cmidrule(lr){2-3}\cmidrule(lr){4-5}\cmidrule(l){6-6}
\rowcolor{tablehead}
\shortstack[l]{\textbf{Source-task}\\\textbf{adaptation}}
& \textbf{MATH}
& \shortstack{\textbf{MATH-}\\\textbf{Hard}}
& \shortstack{\textbf{CruxEval-}\\\textbf{O}}
& \shortstack{\textbf{CruxEval-}\\\textbf{I}}
& \textbf{Avg.} \\
\midrule
\rowcolor{tablegroup}\multicolumn{6}{@{}l}{\textbf{Qwen3-1.7B}} \\
Baseline & \textbf{86.54} & 72.36 & 50.75 & 24.00 & 58.41 \\
FOCUS (Sudoku) & 86.38 & \textbf{72.96} & \textbf{55.38} & \textbf{26.75} & \textbf{60.37} \\
FOCUS (Maze) & 86.36 & 71.68 & 50.63 & 24.12 & 58.20 \\
\addlinespace[2pt]
\rowcolor{tablegroup}\multicolumn{6}{@{}l}{\textbf{Qwen3-4B}} \\
Baseline & 91.58 & 83.46 & 79.63 & 45.25 & 74.98 \\
FOCUS (Sudoku) & 90.92 & 82.40 & 79.50 & 44.25 & 74.27 \\
FOCUS (Maze) & \textbf{92.12} & \textbf{83.91} & \textbf{79.75} & \textbf{45.50} & \textbf{75.32} \\
\addlinespace[2pt]
\rowcolor{tablegroup}\multicolumn{6}{@{}l}{\textbf{Qwen3-8B}} \\
Baseline & 91.14 & 82.48 & 81.63 & \textbf{57.75} & 78.25 \\
FOCUS (Sudoku) & 91.48 & 82.40 & \textbf{82.38} & 55.38 & 77.91 \\
FOCUS (Maze) & \textbf{91.52} & \textbf{85.20} & 82.13 & 54.88 & \textbf{78.43} \\
\addlinespace[2pt]
\rowcolor{tablegroup}\multicolumn{6}{@{}l}{\textbf{Llama-3.2-3B-Instruct}} \\
Baseline & \textbf{47.40} & 21.90 & 15.25 & 0.00 & 21.14 \\
FOCUS (Sudoku) & 45.28 & 21.15 & 15.88 & \textbf{0.38} & 20.67 \\
FOCUS (Maze) & \textbf{47.40} & \textbf{22.43} & \textbf{16.88} & 0.25 & \textbf{21.74} \\
\addlinespace[2pt]
\rowcolor{tablegroup}\multicolumn{6}{@{}l}{\textbf{Llama-3.1-8B-Instruct}} \\
Baseline & 49.16 & 22.66 & \textbf{27.00} & 21.75 & 30.14 \\
FOCUS (Sudoku) & \textbf{49.96} & 22.81 & 26.75 & 21.50 & 30.26 \\
FOCUS (Maze) & 49.86 & \textbf{23.49} & 26.75 & \textbf{22.00} & \textbf{30.53} \\
\bottomrule
\end{tabularx}
\par\smallskip
\begin{minipage}{\textwidth}
\footnotesize
Zero-shot CoT prompting uses no worked examples. Baseline is the unadapted backbone; FOCUS (Sudoku) and FOCUS (Maze) use its language-model LoRA adapter after training on the named source task. Avg.\ is the unweighted mean over the four benchmarks. Bold marks the nonzero maximum within each backbone and column.
\end{minipage}
\end{table}



\label{sec:matched_results}
\label{sec:horizon_analysis}

We study three design choices in FOCUS: how replay states are selected, whether local repair progress is regularized, and whether replay restores the recurrent memory associated with a selected state. Table~\ref{tab:curation_comparison} compares complete curation schemes. The remaining ablations change one component of FOCUS while holding the rest of the training recipe fixed.

\begin{wraptable}{r}{0.62\linewidth}
\centering
\vspace{-1em}
\caption{State-curation comparison with Qwen3-1.7B.
Held-out exact solve accuracy (\%) on Sudoku-Extreme
($K=128$) and Maze-Hard ($K=16$).}\label{tab:curation_comparison}
\label{tab:matched_ablation}
\vspace{-1em}
\footnotesize
\setlength{\tabcolsep}{3pt}
\renewcommand{\arraystretch}{1.12}
\begin{tabularx}{\linewidth}{@{}l>{\raggedright\arraybackslash}XC{1.15cm}C{0.95cm}@{}}
\toprule
\rowcolor{tablehead}
\textbf{Method} & \textbf{Replay-state selection} &
\shortstack{\textbf{Sudoku-}\\\textbf{Extreme}} &
\shortstack{\textbf{Maze-}\\\textbf{Hard}} \\
\midrule
Random & Uniformly sampled rollout state & 57.8 & 89.7 \\
Energy-Hard & Rollout state with the highest task error & 61.3 & 89.9 \\
Fixed Mix & Fixed mixture of training-state types & 60.8 & 90.5 \\
\rowcolor{tablerow}
\textbf{FOCUS} & Rollout state nearest the repair frontier & \textbf{64.4} & \textbf{91.1} \\
\bottomrule
\end{tabularx}
\vspace{-2em}
\parbox{\linewidth}{\footnotesize Bold marks the highest accuracy in each task.}
\end{wraptable}

\paragraph{State-curation comparison.}
Table~\ref{tab:curation_comparison} compares four training-state selection methods with Qwen3-1.7B. Random samples rollout states uniformly. Energy-Hard selects the state with the highest task energy. Fixed Mix samples initial states, corrupted targets, and rollout states in fixed proportions. FOCUS selects the rollout state whose short-horizon energy reduction is closest to the contraction target and applies the defect penalty during replay.

On Sudoku-Extreme, FOCUS achieves $64.4\%$ exact accuracy,
compared with $61.3\%$ for Energy-Hard.
On Maze-Hard, FOCUS achieves $91.1\%$, compared with
$90.5\%$ for Fixed Mix.
We next examine the defect penalty and state-selection rule separately.

\begin{table}[!htbp]
\centering
\caption{\textbf{Defect-penalty ablation.} Qwen3-1.7B exact solve accuracy (\%) with and without the penalty.}
\label{tab:defect_penalty_ablation}
\label{tab:defect_ablation}
\small
\setlength{\tabcolsep}{7pt}
\renewcommand{\arraystretch}{1.14}
\begin{tabular}{@{}lcc@{}}
\toprule
\rowcolor{tablehead}
\textbf{Method} & \textbf{Sudoku-Extreme} & \textbf{Maze-Hard} \\
\midrule
FOCUS w/o defect penalty & 61.3 & 85.8 \\
\rowcolor{tablerow}
FOCUS & \textbf{64.4} & \textbf{91.1} \\
\bottomrule
\end{tabular}
\par\smallskip
{\footnotesize Both variants retain frontier selection and full-state replay. Bold marks the higher accuracy in each task.}
\end{table}

\paragraph{Defect penalty.}
We remove the defect penalty by setting $\beta_{\mathrm{def}}=0$ while retaining frontier selection and full-state replay. All other training and evaluation settings remain unchanged. Table~\ref{tab:defect_ablation} shows that removing the penalty reduces exact accuracy from $64.4\%$ to $61.3\%$ on Sudoku-Extreme and from $91.1\%$ to $85.8\%$ on Maze-Hard.

\paragraph{Frontier versus random selection.}
We replace frontier selection with uniform sampling from the same eligible rollout states. The defect penalty, full-state replay, training budget, and all other training and evaluation settings remain unchanged. On Sudoku-Extreme at $K=128$, replacing frontier selection with uniform sampling reduces exact accuracy from $64.4\%$ to $61.1\%$. On Maze-Hard at $K=16$, accuracy decreases from $91.1\%$ to $87.1\%$. Table~\ref{tab:selection_ablation} in Appendix~\ref{app:selection_ablation} reports both comparisons.

\subsection{Recurrent Repair Trajectories}
\label{sec:repair_trajectories}

\begin{figure}[!htbp]
    \centering
    \includegraphics[width=\linewidth]{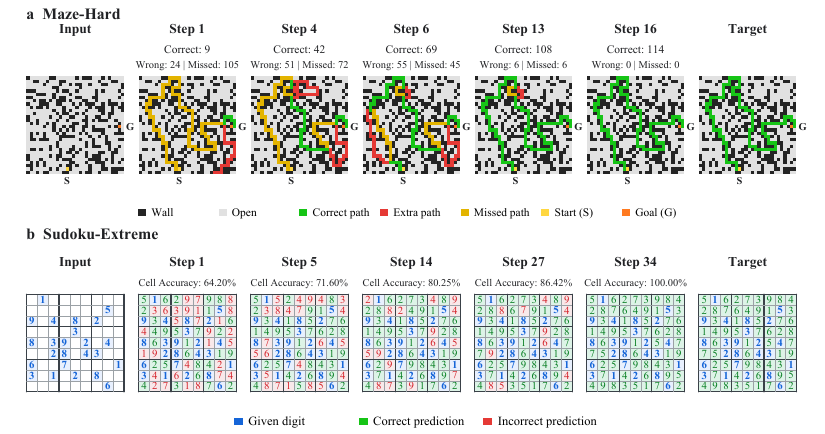}
   \caption{Recurrent repair with Qwen3-1.7B FOCUS.
(a) The Maze-Hard example first matches the target at step 16.
(b) The Sudoku-Extreme example first matches the target at step 34.
Maze path counts exclude the start and goal;
Sudoku cell accuracy includes all 81 cells.}
    \label{fig:repair_trajectories}
\end{figure}

Figure~\ref{fig:repair_trajectories} follows a held-out example from each task. In Maze-Hard, the predicted path first matches the reference at $t=16$. In Sudoku-Extreme, the displayed predictions contain 29, 23, 16, 11, and then zero incorrect cells, with the first exact grid at $t=34$. The rightmost panel in each row is the target. Longer repair trajectories are shown in Appendices~\ref{app:sudoku_repair_trajectory} and~\ref{app:maze_repair_trajectory}.

\begin{figure}[!htbp]
  \centering
  \includegraphics[width=\linewidth,height=0.50\textheight,keepaspectratio]{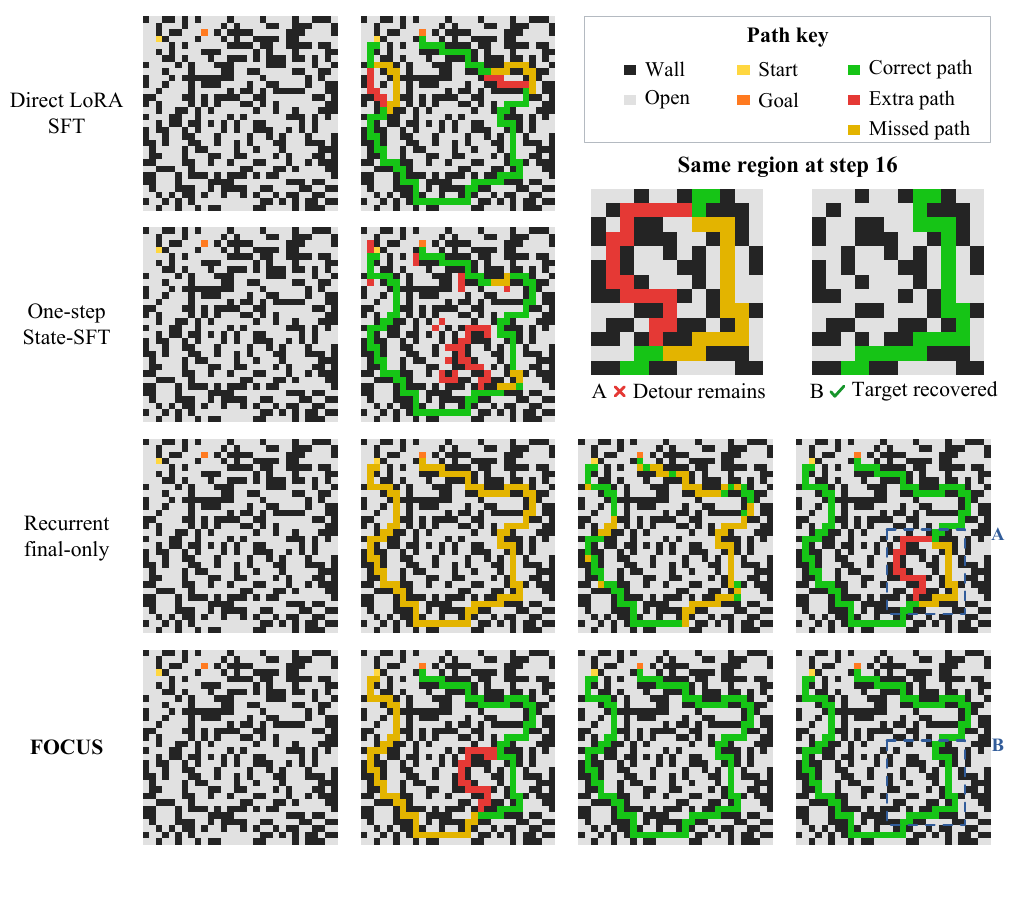}
  \caption{Maze-Hard predictions with Qwen3-8B. Each row starts with the input.
  Direct and One-step show final predictions; Recurrent final-only and FOCUS
  show steps 4, 7, and 16. The recurrent baseline ends with a valid but non-shortest
  path; FOCUS matches the reference at step 7 and retains it at step 16.
  Insets A and B magnify the boxed regions at step 16.
  Colors indicate agreement with the reference after inference.}
  \label{fig:maze_baseline_comparison}
\end{figure}

\paragraph{Comparison with baselines.}
Figure~\ref{fig:maze_baseline_comparison} compares predictions on the same Maze-Hard test instance. Direct LoRA SFT and One-step State-SFT predict extra path cells and omit parts of the reference path. Recurrent final-only progressively forms a valid route but retains a detour at step 16. FOCUS removes the extra path cells and recovers the missing segments, matching the shortest reference path at step 7 and preserving it through step 16.
Appendix~\ref{app:qualitative_baseline_comparisons} provides the corresponding Sudoku-Extreme comparison.

\section{Conclusion}

We studied whether pretrained language models can support iterative explicit-state repair and how the states used to train such a solver should be chosen. Coupling an adapted language model with a shared recurrent updater improves exact solve accuracy over direct prediction and a single update on Sudoku-Extreme and Maze-Hard. FOCUS selects states from the model's own rollouts according to short-horizon repair progress and resumes training from both the saved answer and memory. Ablations on Sudoku-Extreme and Maze-Hard show gains from frontier selection and the defect penalty. With the recurrent updater disabled, the adapted language models also improve zero-shot accuracy on several mathematical and code-execution benchmarks.

\clearpage
\section*{Reproducibility Statement}
The supplementary material provides the implementation, configuration files, and scripts used to train and evaluate FOCUS and the reported baselines. Section~\ref{sec:method}, Appendix~\ref{app:implementation}, and Table~\ref{tab:focus_training_config} specify the architecture, objective, state-selection procedure, optimization settings, and task-specific schedules. Section~\ref{sec:experiments} and Appendix~\ref{app:evaluation} define the data splits, validation-based model selection rules, deterministic decoding procedures, and exact-solve metrics. Appendix~\ref{app:downstream} documents the zero-shot prompts and downstream evaluation protocol. The experiments use publicly available pretrained Qwen and Llama backbones and public downstream benchmarks. We provide the fixed held-out evaluation splits, random seeds, checkpoints, and per-instance predictions needed to reproduce the reported source-task and transfer results.

\section*{AI Use Statement}
In preparing this paper, we used Large Language Models only to aid with language polishing, grammar refinement, and improving readability. All research ideas, methodological contributions, experimental design, and analysis were conceived and carried out entirely by the authors. The LLM
was not involved in ideation, technical writing of results, or scientific claims. The authors take full
responsibility for all content of this paper.

\FloatBarrier
\bibliography{references}
\bibliographystyle{iclr2027_conference}

\clearpage
\appendix
\raggedbottom
\section{Proof of the Finite-Horizon Energy Bound}
\label{app:proof}

We use the input-conditioned transition $F_{\Theta,x}$ defined in Section~\ref{sec:recurrent_updater}.

Let $s_{b+1}=F_{\Theta,x}^{h}(s_b)$ for $b=0,\ldots,B-1$, and write $E_b=E_x(\pi_z s_b)$. From \eqref{eq:defect},
\begin{equation}
D_b=\log(E_{b+1}+\epsilon)-\log(E_b+\epsilon)-h\log\rho.
\end{equation}
Summing over blocks telescopes:
\begin{equation}
\log(E_B+\epsilon)=\log(E_0+\epsilon)+Bh\log\rho+\sum_{b=0}^{B-1}D_b.
\end{equation}
Since $D_b\leq[D_b]_+$, exponentiation yields the finite-horizon bound, with $K=Bh$:
\begin{equation}
E_x(z_K)+\epsilon
\leq
\left(E_x(z_0)+\epsilon\right)\rho^K
\exp\!\left(\sum_{b=0}^{B-1}\relu{D_b}\right).
\label{eq:energy_bound}
\end{equation}
The bound follows from an exact accounting identity for task-energy change; it is not a global convergence theorem for the neural transition.

For a tolerance \(\delta>0\), a conceptual frontier band can be written as
\[
\mathcal F_{\theta,\gamma,\delta}(x)
=\{s:|D_\theta^{(h,\rho)}(x,s)-\gamma|\leq\delta\}.
\]
The implemented selector minimizes distance to \(\gamma\) over eligible
candidates. The tolerance describes the band; it does not introduce an
additional acceptance gate or alter the selected candidate.

\section{Task-Energy Definitions}
\label{app:energy}

\subsection{Sudoku}
Let $p_{i,d}=\operatorname{softmax}(z_i)_d$. A practical energy is
\begin{equation}
E_x^{\mathrm{Sudoku}}(z)
=\lambda_{\mathrm{CE}}\ell_{\mathrm{CE}}(p,y^\star)
+\lambda_{\mathrm{hard}}\ell_{\mathrm{hard}}(p,y^\star)
+\lambda_c V_{\mathrm{allDiff}}(p),
\end{equation}
where
\begin{equation}
V_{\mathrm{allDiff}}(p)
=\frac{1}{27\cdot9}\sum_{g\in\mathcal{G}}\sum_{d=1}^{9}
\left(\sum_{i\in g}p_{i,d}-1\right)^2,
\end{equation}
with $\mathcal{G}$ containing nine rows, nine columns, and nine boxes. The hard-cell term is the mean cross-entropy over the highest-loss fraction of non-given cells. Both cross-entropy terms are restricted to non-given cells, and givens are clamped after every update.

\subsection{Maze-Hard}
Let $p_i$ be the probability that open cell $i$ belongs to the path. A practical energy is
\begin{equation}
E_x^{\mathrm{Maze}}(z)
=\ell_{\mathrm{CE}}^{\mathrm{open}}(p,y^\star)
+\lambda_{\mathrm{dice}}\ell_{\mathrm{Dice}}(p,y^\star)
+\lambda_{\mathrm{rank}}V_{\mathrm{rank}}(p)
+\lambda_{\mathrm{dec}}V_{\mathrm{dec}}(p).
\end{equation}
Here $\ell_{\mathrm{Dice}}$ is a soft Dice loss~\citep{milletari2016vnet}.
Here $V_{\mathrm{rank}}$ is a top-$k$ path/non-path boundary-margin penalty, and $V_{\mathrm{dec}}$ is a differentiable surrogate for the fixed decoder. Its task-specific components and weights are specified in the released configuration files. Exact path validity is computed only after decoding and is not fed back during evaluation.

\section{Implementation Details}
\label{app:implementation}

This appendix describes training for the recurrent explicit-state solver.
The four state-curation methods in Table~\ref{tab:curation_comparison},
Random, Energy-Hard, Fixed Mix, and FOCUS, share the recurrent architecture
but differ in how training states are constructed or selected.
FOCUS additionally uses the finite-horizon defect penalty.

\subsection{FOCUS Candidate-State Construction}
\label{app:focus_candidates}

FOCUS constructs additional training states from a rollout produced by the
current model. For an input \(x\), let
\[
s_t=(z_t,u_t),
\]
where \(z_t\) is the explicit answer state and \(u_t\) is the recurrent
memory. Starting from the task-defined initial state \(s^{\mathrm{clean}}_0\),
we generate a short selection rollout
\[
\tau_x=
\left(
s_0,s_1,\ldots,s_{K_p}
\right)
\]
using the current recurrent solver. This rollout is evaluated without gradient,
and complete states \((z_t,u_t)\) are stored so that replay resumes from an
actually visited recurrent state rather than reconstructing either component
independently.

Eligible update indices are restricted to
\[
\mathcal{C}_x
\subseteq
\left\{
t\in\mathcal{T}:t+h\le K_p
\right\},
\]
which guarantees that both frontier scoring and the subsequent local
\(h\)-step training rollout remain within the prespecified selection-rollout horizon.
For every eligible candidate,
\[
D_t
=
D_{\theta}^{(h,\rho)}(x,s_t)
\]
is evaluated without gradient. FOCUS selects
\[
t^\star
=
\arg\min_{t\in\mathcal{C}_x}
\left|
D_t-\gamma
\right|.
\]
If no candidate satisfies the task-specific eligibility conditions,
Sudoku-Extreme omits the auxiliary replay loss while retaining its base
loss. Maze-Hard instead uses an update from the task-defined initial state.

\paragraph{Sudoku-Extreme.}
The explicit answer state contains digit logits
\[
z_t\in\mathbb{R}^{81\times 9}.
\]
Given cells are clamped to their observed values after every recurrent
update. The replay update is omitted when the short-rollout endpoint already
decodes to the reference solution; otherwise the scheduled eligible indices
are used. Exactness is computed with the reference target only to determine
the training update and is unavailable to the inference procedure.

\paragraph{Maze-Hard.}
For Maze-Hard, the answer state contains logits indicating whether each cell belongs to the path,
\[
z_t^{\mathrm{path}}\in\mathbb{R}^{900\times 2},
\]
together with the decoder-aligned state variables used by the deterministic
canonical-path decoder. We denote the complete answer state by
\[
z_t=
\left(
z_t^{\mathrm{path}},
z_t^{\mathrm{dec}}
\right).
\]
The recurrent memory \(u_t\) is replayed together with the complete answer
state.

A candidate is eligible when its deterministic decoded path is non-exact,
or when it is exact at time \(t\) but becomes non-exact after the
\(h\)-step continuation. The latter condition retains exact states
whose subsequent recurrent evolution leaves the target solution; FOCUS then
applies the same finite-horizon frontier score to all eligible states.
Exactness is used only during training for candidate eligibility and is not
used for inference-time search, verification, or repair.

\subsection{Fixed Mix}
\label{app:fixed_mix}

Fixed Mix uses the same recurrent solver but replaces model-adaptive frontier
selection with a predetermined mixture over training-state types. Let
\(q_{\mathrm{clean}}\), \(q_{\mathrm{corrupt}}\), and
\(q_{\mathrm{self}}\) denote task-defined initial states, controlled
corruptions of target states, and states visited during model rollouts. Fixed Mix samples
\[
q_{\mathrm{FM}}(s\mid x,y^\star)
=
\pi_{\mathrm{clean}}q_{\mathrm{clean}}
+
\pi_{\mathrm{corrupt}}q_{\mathrm{corrupt}}
+
\pi_{\mathrm{self}}q_{\mathrm{self}}.
\]
The initial-state/corruption/rollout probabilities are
\(0.50/0.25/0.25\) for Sudoku and \(0.45/0.35/0.20\) for Maze-Hard.
It does not compute the repair-frontier
score.

\subsection{Task-Specific Training Configuration}
\label{app:training_config}

Table~\ref{tab:implementation_config} summarizes the Qwen3-1.7B FOCUS
training settings for Sudoku-Extreme and Maze-Hard.
We use AdamW~\citep{loshchilov2019decoupled}.
Fixed Mix sampling is defined in Appendix~\ref{app:fixed_mix};
differences between the compared training schemes are described in
Appendix~\ref{app:baseline_definitions}.

\begin{table}[!htbp]
\centering
\caption{Training settings for Qwen3-1.7B FOCUS on Sudoku-Extreme and Maze-Hard.}
\label{tab:implementation_config}\label{tab:focus_training_config}
\footnotesize
\setlength{\tabcolsep}{5pt}
\renewcommand{\arraystretch}{1.18}
\begin{tabularx}{\textwidth}{@{}L{2.6cm}>{\raggedright\arraybackslash}X>{\raggedright\arraybackslash}X@{}}
\toprule
\rowcolor{tablehead}
\textbf{Setting}
& \textbf{Sudoku-Extreme}
& \textbf{Maze-Hard} \\
\midrule
\multicolumn{3}{@{}l}{\textit{Model and optimization}} \\
\addlinespace[2pt]

Optimizer
& AdamW; weight decay \(0.01\)
& AdamW; weight decay \(0.01\) \\

Training updates
& \(50{,}000\)
& \(16{,}000\) \\

Effective batch size
& \(16\) (mini-batch \(2\), accumulation \(8\))
& \(16\) (mini-batch \(1\), accumulation \(16\)) \\

Recurrent updater
& Hidden size 256; 2 layers; 4 attention heads; dropout \(0.05\)
& Hidden size 256; 2 layers; 4 attention heads; dropout \(0.05\) \\

Answer-update scale
& \(0.80\) (multiplier on the predicted logit increment)
& \(0.50\) (multiplier on the predicted logit increment) \\

LoRA
& \(r=16,\;\alpha=32\), dropout \(0.05\);
  attention and MLP projections
& \(r=16,\;\alpha=32\), dropout \(0.05\);
  attention and MLP projections \\

Learning-rate schedule
& Linear warmup followed by half-cosine interpolation through fixed knots;
  recurrent-updater peak \(3\times10^{-4}\);
  LoRA peak \(10^{-5}\)
& 300-step warmup followed by cosine decay;
  recurrent-updater peak \(2\times10^{-4}\);
  LoRA peak \(10^{-5}\) \\

\midrule
\multicolumn{3}{@{}l}{\textit{State selection and replay}} \\
\addlinespace[2pt]

Rollout \(K_p\); replay \(h\)
& \((K_p,h)=(16,4),(32,4),(64,8),\allowbreak(96,8),(128,8)\)
& \(K_p=16;\;h=4\) \\

Candidate rollout steps
& \(0,2,4,8,12,16,24,32,\allowbreak48,64,80,96,112\),
  restricted by \(t+h\le K_p\)
& \(0,2,4,6,8,10,12\) \\

Replay schedule
& Starts at update \(3{,}000\);
  scheduled fraction of mini-batches with auxiliary replay:
  \(0.175\rightarrow0.325\);
  auxiliary-loss weight \(w_{\mathrm{aux}}=0.45\)
& Starts at update \(3{,}000\);
  auxiliary-replay probability:
  \(0.12\rightarrow0.30\) over \(4{,}500\) updates \\

\midrule
\multicolumn{3}{@{}l}{\textit{Training objectives}} \\
\addlinespace[2pt]

Defect parameters
\((\rho,\gamma,\epsilon)\);
\(\beta_{\mathrm{def}}\)
& \((0.985,0,0.05);\;0.08\)
& \((0.985,0,0.05);\;0.05\) \\

Initial-state supervision
& Task loss from the initialization in every mini-batch
& Initial-state sampling probability \(0.45\) when auxiliary replay
  is not scheduled; overall probability at least \(0.315\) \\

Task-energy coefficients
& \((\lambda_{\mathrm{CE}},\lambda_{\mathrm{hard}},\lambda_c)
  =(1.00,0.25,0.20)\)
& Open-cell cross-entropy coefficient \(1.00\);
  \((\lambda_{\mathrm{dice}},\lambda_{\mathrm{rank}},\lambda_{\mathrm{dec}})
  =(0.40,0.04,0.10)\) \\

Teacher supervision from the initial state
& Frozen 16-step TRM;
  KL-loss weight \(0.05\);
  temperature \(2\);
  starts at update \(800\) with a \(1{,}200\)-step warmup;
  at most 4 student steps
& Cached outputs from a frozen 16-step TRM;
  KL-loss weight \(0.005\);
  temperature \(4\);
  starts at update \(3{,}000\) with a \(2{,}000\)-step warmup;
  at most 4 student steps \\

\bottomrule
\end{tabularx}
\end{table}

The accompanying configuration files provide the complete learning-rate
and replay schedules.

\subsection{TRM Teacher Supervision}
\label{app:teacher}

A frozen TRM teacher provides an additional temperature-scaled
KL-divergence loss for training from the task-defined initial state.
Teacher supervision is not applied to auxiliary replay and is not used
during state selection or inference.
Table~\ref{tab:implementation_config} lists the task-specific teacher settings.

\subsection{Training Algorithm}
\label{app:focus_algorithm}

Algorithm~\ref{alg:focus} summarizes the FOCUS training procedure.
The ordinary training loss samples from \(q_{\mathrm{base}}\)
(Appendix~\ref{app:baseline_definitions}) and includes teacher supervision
only for initial-state training (Appendix~\ref{app:teacher}).

\begin{algorithm}[!htbp]
\caption{FOCUS training}
\label{alg:focus}
\small
\begin{algorithmic}[1]
\REQUIRE Data \(\mathcal D\); trainable parameters \(\Theta=(\theta,\phi,\psi)\);
training updates \(U\); rollout horizon \(K_p\); replay horizon \(h\)
\ENSURE Trained parameters \(\Theta_U\)
\FOR{\(n=1,\ldots,U\)}
    \STATE \(\mathcal L\gets0\)
    \FOR{each accumulated mini-batch \(B_m\)}
        \STATE Sample \(B_m\) from \(\mathcal D\); draw \(a\sim\mathrm{Bernoulli}(p_{n,m})\)
        \FORALL{\((x,y^\star)\in B_m\)}
            \STATE Compute \(\ell_{\mathrm{base}}\) if Sudoku or \(a=0\); otherwise set \(\ell_{\mathrm{base}}\gets0\)
            \STATE \(\ell_{\mathrm{aux}}\gets0\)
            \IF{\(a=1\)}
                \STATE Generate \(\tau=(s_0^{\mathrm{clean}},\ldots,s_{K_p})\) without gradients
                \STATE Construct \(\mathcal C_x\) as in Appendix~\ref{app:focus_candidates};
                evaluate \(D_\theta^{(h,\rho)}(x,s_t)\) without gradients for \(t\in\mathcal C_x\)
                \IF{\(\mathcal C_x\neq\varnothing\)}
                    \STATE \(t^\star\gets\arg\min_{t\in\mathcal C_x}
                    \left|D_\theta^{(h,\rho)}(x,s_t)-\gamma\right|\)
                    \STATE \(s_f\gets\operatorname{stopgrad}(s_{t^\star})\), retaining both \(z_{t^\star}\) and \(u_{t^\star}\)
                    \STATE Recompute \(R_{\phi,\psi}(x)\) with gradients
                    \STATE Replay \(h\) updates with gradients from \(s_f\), using \(R_{\phi,\psi}(x)\) and \(c_x\)
                    \STATE Evaluate \(\mathcal L_{\mathrm{task}}\) and \(D_\theta^{(h,\rho)}(x,s_f)\) on this replay
                    \STATE \(\ell_{\mathrm{aux}}\gets\mathcal L_{\mathrm{task}}
                    +\beta_{\mathrm{def}}\left[D_\theta^{(h,\rho)}(x,s_f)-\gamma\right]^2_+\)
                \ELSE
                    \STATE For Maze-Hard, compute \(\ell_{\mathrm{base}}\); for Sudoku, retain its existing loss
                \ENDIF
            \ENDIF
            \STATE \(\mathcal L\gets\mathcal L+\ell_{\mathrm{base}}+
            w_{\mathrm{aux}}^{\mathrm{task}}\ell_{\mathrm{aux}}\)
        \ENDFOR
    \ENDFOR
    \STATE Update \(\Theta\) using \(\nabla_\Theta\bigl(\mathcal L/\sum_m|B_m|\bigr)\)
\ENDFOR
\end{algorithmic}
\end{algorithm}
\FloatBarrier

Replay probabilities, candidate steps, and loss coefficients are listed
in Table~\ref{tab:implementation_config}.


\section{Evaluation Protocols}
\label{app:evaluation}

\subsection{Source-Task Evaluation}
\label{app:source_eval}
\label{app:backbone_portability}

Exact solve rate is the primary evaluation metric for Sudoku-Extreme and
Maze-Hard.

\paragraph{Sudoku-Extreme.}
Each instance consists of a partially observed \(9\times9\) Sudoku grid
with a unique reference solution. Given digits are immutable: they are
clamped to their observed values after every explicit-state update.

A Sudoku instance is counted as exactly solved only when the final decoded
grid matches the unique reference solution in all \(81\) cells:
\[
\operatorname{Exact}_{\mathrm{Sudoku}}(\hat y,y^\star)
=
\mathbb{I}
\left[
\hat y_i=y_i^\star
\;\;\forall i\in\{1,\ldots,81\}
\right].
\]
Exact solve rate is the mean of this indicator over the evaluation set.

For the trajectory visualizations, cell accuracy measures agreement with
the reference solution over all \(81\) cells, including givens.

The Sudoku-Extreme corpus contains \(422{,}786\) puzzles.
We use its first \(1{,}000\) puzzles for validation and sample
\(1{,}000\) test puzzles uniformly without replacement from the
remaining puzzles using seed \(20260831\).
Training uses \(71{,}407\) symmetry-augmented examples and
\(50{,}000\) synthetic puzzles.
We verified that neither training source overlaps with the validation
or test puzzles. All compared methods use the same test set.

\paragraph{Maze-Hard.}
Maze-Hard uses fixed train/validation/test splits of
\(900/100/1{,}000\) instances. Each maze contains a start cell \(S\), a
goal cell \(G\), traversable cells, and walls.

A breadth-first search with fixed
down--left--right--up neighbor expansion order and first-discovery parent
assignment defines a unique canonical shortest-path target. This
canonicalization makes exact solve well-defined even when multiple
shortest paths exist geometrically.

FOCUS, Random, Energy-Hard, and Fixed Mix use a deterministic
canonical-path decoder based on predicted parent probabilities and
distances to the goal.
Starting from the goal, the decoder repeatedly selects a traversable
neighbor
\[
j^\star
=
\arg\max_{j\in\mathcal{N}(i)}
\left[
\log p_{\mathrm{parent}}(j\mid i)
-
0.35
\left|
d_i-d_j-1
\right|
\right],
\]
where \(i\) is the current cell, \(\mathcal{N}(i)\) contains its
traversable neighbors, \(p_{\mathrm{parent}}\) is the learned parent
probability, and \(d\) is the learned distance prediction. Exact score
ties are resolved using the same fixed down--left--right--up order.

The decoder follows at most \(256\) path transitions and stops earlier
if it reaches the start cell or encounters a loop.
An unsuccessful walk remains partial.

A Maze-Hard instance is counted as exactly solved only when the
deterministically decoded path matches the protocol-defined canonical
shortest-path target exactly:
\[
\operatorname{Exact}_{\mathrm{Maze}}(\hat y,y^\star)
=
\mathbb{I}
\left[
\hat y=y^\star
\right].
\]

Diagnostic metrics include path validity, shortest-path optimality, and
path F1. A valid path connects the start and goal through traversable
cells as a single non-branching route; it is optimal if its length equals
the shortest-path length. Path F1 is computed over traversable cells,
excluding the two endpoints, and averaged across mazes.
A valid shortest path need not match the canonical reference when
multiple shortest paths exist.

The Maze One-step State-SFT and Recurrent final-only baselines instead classify path cells using their prespecified threshold; they do not require the canonical-path prediction heads. All methods are scored against the same canonical reference.

\paragraph{Prompt-only reasoning models.}
Table~\ref{tab:source_task_prompt_only} reproduces the results reported by
\citet{wang2025hrm} for DeepSeek-R1, Claude 3.7 8K, and o3-mini (high).
All three obtain \(0\%\) exact solve accuracy on Sudoku-Extreme and
Maze-Hard, meaning that no evaluated instance receives a fully correct
solution. Partial grids and incomplete paths do not count as solved
instances. Both tasks require coordinating decisions across the entire
grid, where a locally plausible digit or path segment can conflict with
the remaining solution. This motivates maintaining an explicit candidate
solution that can be revised across successive updates.

\paragraph{Model selection.}
Checkpoints are selected by exact solve rate on the validation set.
The state-curation comparison uses \(K=128\) on Sudoku-Extreme and
\(K=16\) on Maze-Hard. For the depth analysis, \(K\) is also selected
on the validation set from the candidate horizons reported in the main
text, with ties favoring smaller \(K\).
The test sets are used only for final evaluation.

\paragraph{Inference protocols.}
The number of recurrent updates depends on the method.

\begin{itemize}
    \item \textbf{Direct prediction.}
    Direct models predict the structured output without recurrent updates.
    A fixed task parser or deterministic decoder produces the final answer.
    Malformed outputs are counted as incorrect.

    \item \textbf{One-step State-SFT.}
    One-step models initialize the task-defined state, apply
    exactly one learned state update, and decode once.

    \item \textbf{Recurrent models.}
    Recurrent final-only, Fixed Mix, and FOCUS models initialize the same
    task-defined initial state, apply exactly \(K\) updates of the shared
    recurrent transition, and decode once.
\end{itemize}

Each method produces one final prediction.
Ground-truth targets, teacher outputs, and training-time selection scores
are unavailable during prediction; reference targets are used only for
scoring afterward. No additional search, verifier-guided correction,
parameter optimization, or retries are used.

\subsection{Baseline Definitions}
\label{app:baselines}\label{app:baseline_definitions}

We compare direct prediction, one-step state prediction, recurrent
prediction, and training with additional replay states.
All methods use the same task inputs and exact-match metric.
Evaluation follows Appendix~\ref{app:source_eval}.

\paragraph{Base.}
The unadapted pretrained or instruction-tuned backbone produces a single
greedy prediction from the textual puzzle input.

\paragraph{Direct LoRA SFT.}
Only the language-model LoRA parameters are trained to predict the target
directly. For Sudoku-Extreme, the output contains the \(81\) digits of
the reference solution.

\paragraph{One-step State-SFT.}
The model applies one learned update from the task-defined initial state,
\[
s_1=F_\theta\left(s_0^{\mathrm{clean}};R_{\phi,\psi}(x),c_x\right),
\]
and trains the resulting prediction against the reference target.

\paragraph{Recurrent final-only.}
The model repeatedly applies the shared updater from the task-defined
initial state,
\[
s_{t+1}=F_\theta\left(s_t;R_{\phi,\psi}(x),c_x\right),
\qquad t=0,\ldots,K-1.
\]
Task supervision is applied only to the final prediction at \(s_K\).

\paragraph{Fixed Mix.}
Additional training states are sampled from the mixture defined in
Appendix~\ref{app:fixed_mix}. This method does not use the defect penalty.

\paragraph{FOCUS.}
FOCUS uses the state-selection and replay procedure defined in
Section~\ref{sec:method} and Algorithm~\ref{alg:focus}.

\subsection{Defect-Penalty and State-Selection Ablations}
\label{app:curation_ablation}
\label{app:defect_ablation}

We remove the defect penalty by setting \(\beta_{\mathrm{def}}=0\),
while retaining frontier selection and full-state replay.
Within each task, both variants use seed 42 and the same training and
evaluation settings, differing only in the defect penalty.
Evaluation uses \(K=128\) on Sudoku-Extreme and \(K=16\) on Maze-Hard.

Removing the penalty reduces exact solve accuracy from \(64.4\%\)
to \(61.3\%\) on Sudoku-Extreme and from \(91.1\%\) to \(85.8\%\)
on Maze-Hard (Table~\ref{tab:defect_ablation}).

\paragraph{Frontier versus random selection.}
\label{app:selection_ablation}
We replace frontier selection with uniform sampling from the same eligible
rollout states. Both variants retain the defect penalty and full-state replay.
Within each task, training uses seed 42 with the same training budget and
evaluation protocol. Checkpoints are selected on the validation set.
Table~\ref{tab:selection_ablation} reports held-out exact solve accuracy.

\begin{table}[!htbp]
\centering
\caption{\textbf{State-selection ablation.} Qwen3-1.7B exact solve accuracy (\%) on Sudoku-Extreme ($K=128$) and Maze-Hard ($K=16$).}
\label{tab:selection_ablation}
\small
\setlength{\tabcolsep}{7pt}
\renewcommand{\arraystretch}{1.14}
\begin{tabular}{@{}lcc@{}}
\toprule
\rowcolor{tablehead}
\textbf{Method} & \textbf{Sudoku-Extreme} & \textbf{Maze-Hard} \\
\midrule
FOCUS with random selection & 61.1 & 87.1 \\
\rowcolor{tablerow}
FOCUS & \textbf{64.4} & \textbf{91.1} \\
\bottomrule
\end{tabular}
\par\smallskip
{\footnotesize Both variants retain the defect penalty and full-state replay. Bold marks the higher accuracy in each task.}
\end{table}

\subsection{Downstream Benchmarks}
\label{app:downstream_benchmarks}
\label{app:downstream}

We evaluate transfer to mathematical reasoning and code execution using
MATH~\citep{hendrycks2021math}, MATH-Hard, AIME 2025, AIME 2026, and
CruxEval-O/I~\citep{gu2024cruxeval}.

\paragraph{Mathematical reasoning.}
MATH contains \(5{,}000\) test problems; MATH-Hard contains the
\(1{,}324\) Level-5 problems from the same test
split~\citep{lighteval2024mathhard}.
AIME 2025 and AIME 2026 each contain \(30\) problems.
We use the \texttt{test} split of
\texttt{test-time-compute/aime\_2025}~\citep{testtimecompute2025aime}
and the \texttt{train} split of
\texttt{MathArena/aime\_2026}~\citep{dekoninck2026matharena},
respectively. Both are used exclusively for evaluation.

\paragraph{Code execution.}
CruxEval-O evaluates program-output prediction, while CruxEval-I
evaluates input prediction given a program and its output.
Each contains \(800\) examples.

\paragraph{Models and evaluation.}
We evaluate Qwen3-1.7B, Qwen3-4B, Qwen3-8B,
Llama-3.2-3B-Instruct, and Llama-3.1-8B-Instruct.
For each backbone, Base denotes the unadapted model;
FOCUS (Sudoku) and FOCUS (Maze) use the LoRA adapters trained on
Sudoku-Extreme and Maze-Hard, respectively.
The recurrent updater is disabled during downstream evaluation.

Each item is evaluated with a fixed zero-shot chain-of-thought prompt
and one greedy completion.
Within each backbone, the compared variants use the same prompt,
tokenizer, chat template, generation limit, answer extraction, and
scoring procedure.
Source-task checkpoints are selected on the source-task validation
set before downstream evaluation.
No downstream benchmark is used for parameter training or adapter
selection, and adapters are applied with scaling coefficient one.

Mathematical answers are extracted and scored with the
benchmark-specific answer verifier.
CruxEval-O and CruxEval-I use their official evaluators.
Malformed completions follow the same scoring rules for all variants.
No retries or verifier-guided correction are used.

\clearpage
\section{Recurrent Repair on Sudoku-Extreme}
\label{app:sudoku_repair_trajectory}

\begin{figure}[!htbp]
  \centering
  \includegraphics[width=\linewidth]{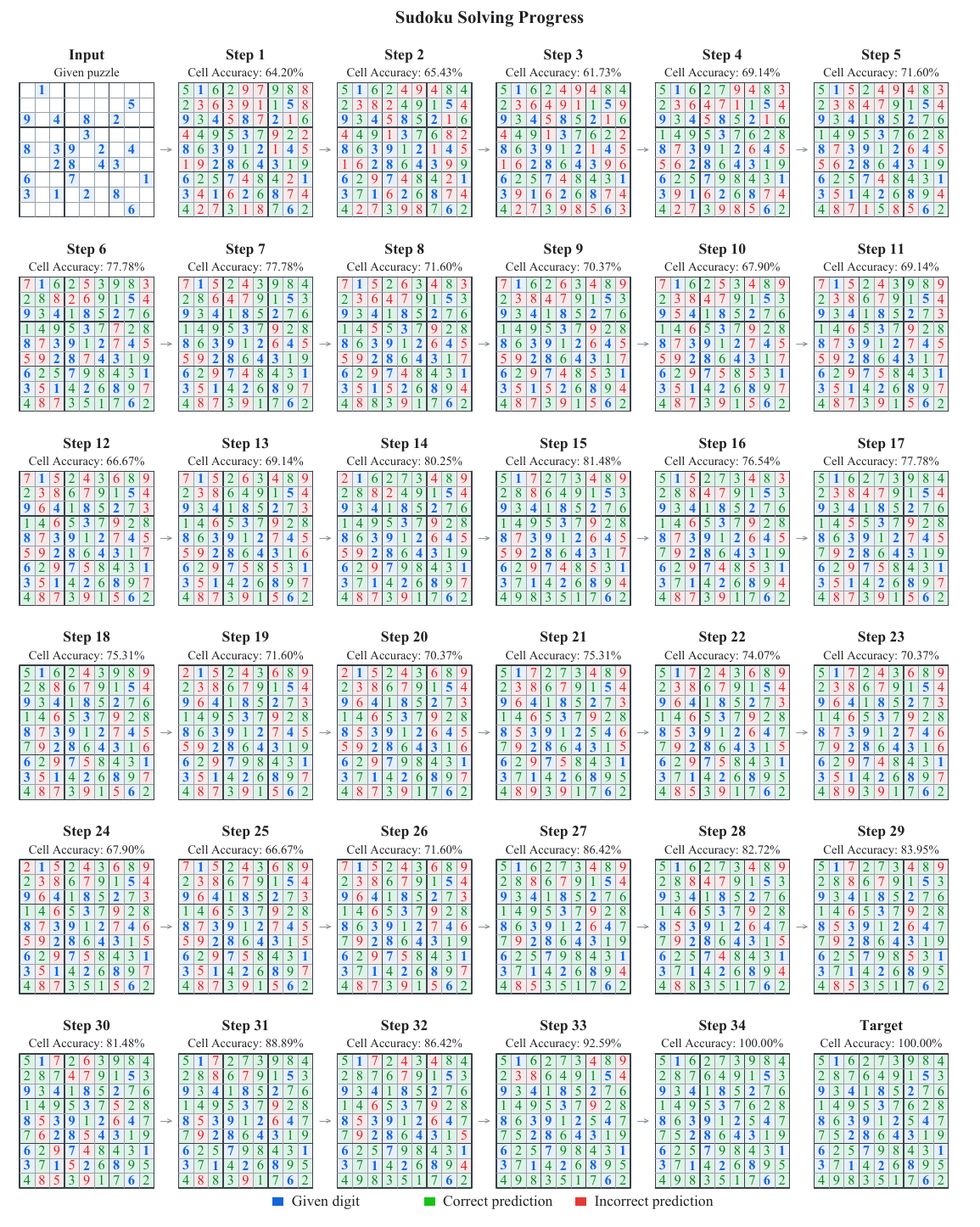}
  \caption{Visualization of a Sudoku puzzle and its solving steps using FOCUS.}
  \label{fig:sudoku_locked_gradual_repair}
\end{figure}

Figure~\ref{fig:sudoku_locked_gradual_repair} shows the first 34 recurrent updates of Qwen3-1.7B
FOCUS on one Sudoku-Extreme test puzzle, with the first
correct solution reached at step 34.

\clearpage
\section{Recurrent Repair on Maze-Hard}
\label{app:maze_repair_trajectory}

\begin{figure}[!htbp]
  \centering
  \includegraphics[width=\linewidth,height=0.75\textheight,keepaspectratio]{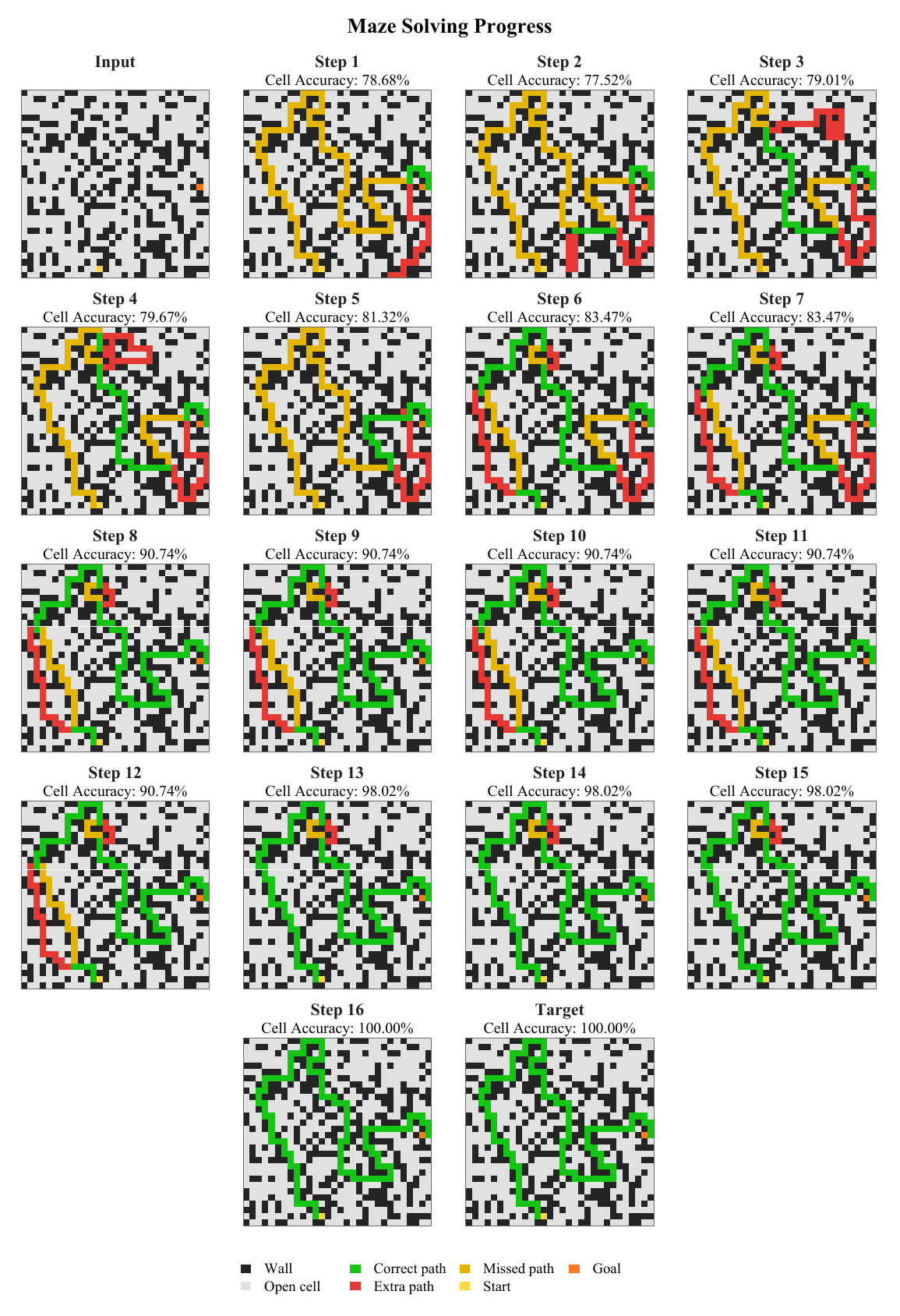}
  \caption{Visualization of a maze and its solving steps using FOCUS.}
  \label{fig:maze_locked_gradual_repair}
\end{figure}

Figure~\ref{fig:maze_locked_gradual_repair} shows Qwen3-1.7B FOCUS solving a Maze-Hard test
instance at step 16. Cell accuracy covers non-wall cells.
The model shown achieves 90.6\% accuracy at $K=16$ and uses a different
model snapshot from the one yielding 91.1\% in Table~\ref{tab:curation_comparison}.

\clearpage
\section{Qualitative Comparisons with Baselines}
\label{app:qualitative_baseline_comparisons}

\begin{figure}[!htbp]
  \centering
  \includegraphics[width=\linewidth]{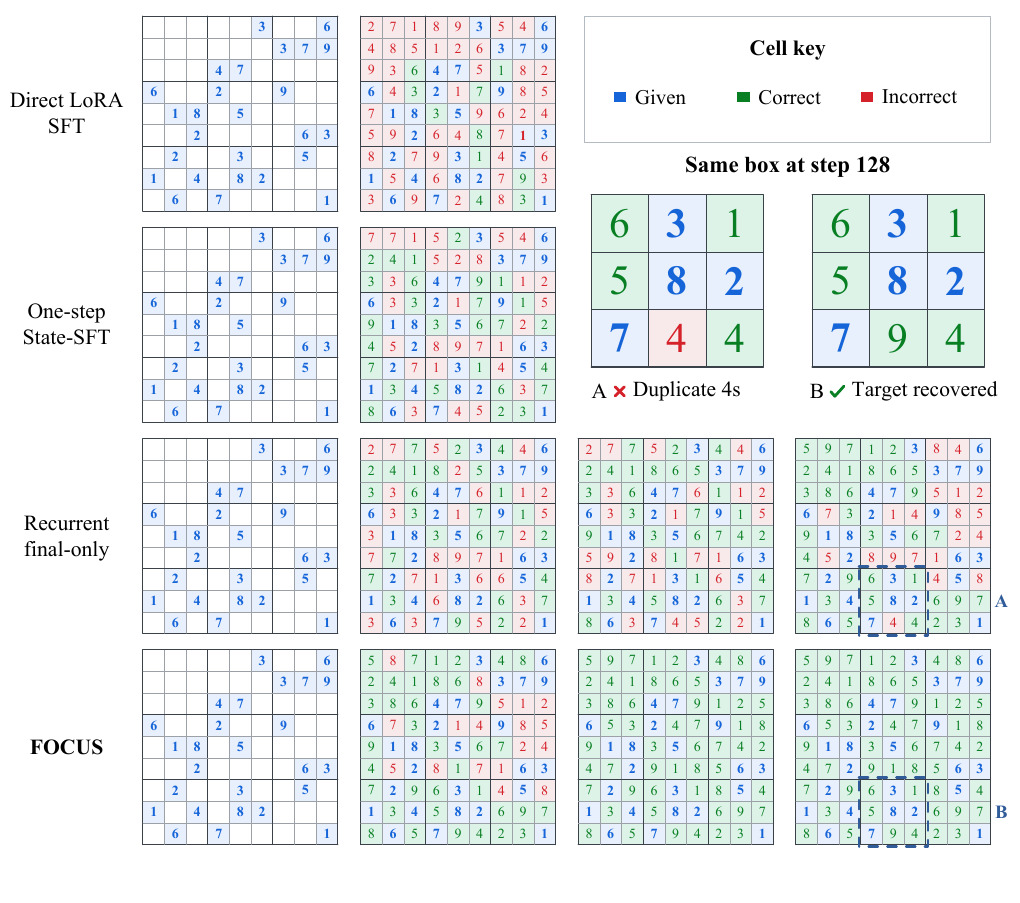}
  \caption{Predictions on the same Sudoku-Extreme test puzzle using Qwen3-8B.
  Each row starts with the input. Direct LoRA SFT and One-step State-SFT show
  their final answers in the second position; Recurrent final-only and FOCUS
  show predictions at steps 4, 8, and 128. FOCUS solves this example at step 8
  and retains the solution through step 128. Insets A and B magnify the boxed regions at step 128.
  Colors indicate agreement with
  the target after inference.}
  \label{fig:sudoku_baseline_comparison}
\end{figure}

Figure~\ref{fig:sudoku_baseline_comparison} compares FOCUS and the baselines on the same Sudoku-Extreme puzzle.

The corresponding Maze-Hard comparison is shown in
Figure~\ref{fig:maze_baseline_comparison} in the main text.

\clearpage
\section{Qualitative Examples of Downstream Transfer}
\label{app:downstream_qualitative}

\begin{figure}[!htbp]
  \centering
  \includegraphics[page=1,width=\linewidth,height=0.85\textheight,keepaspectratio,trim=24bp 259bp 24bp 24bp,clip]{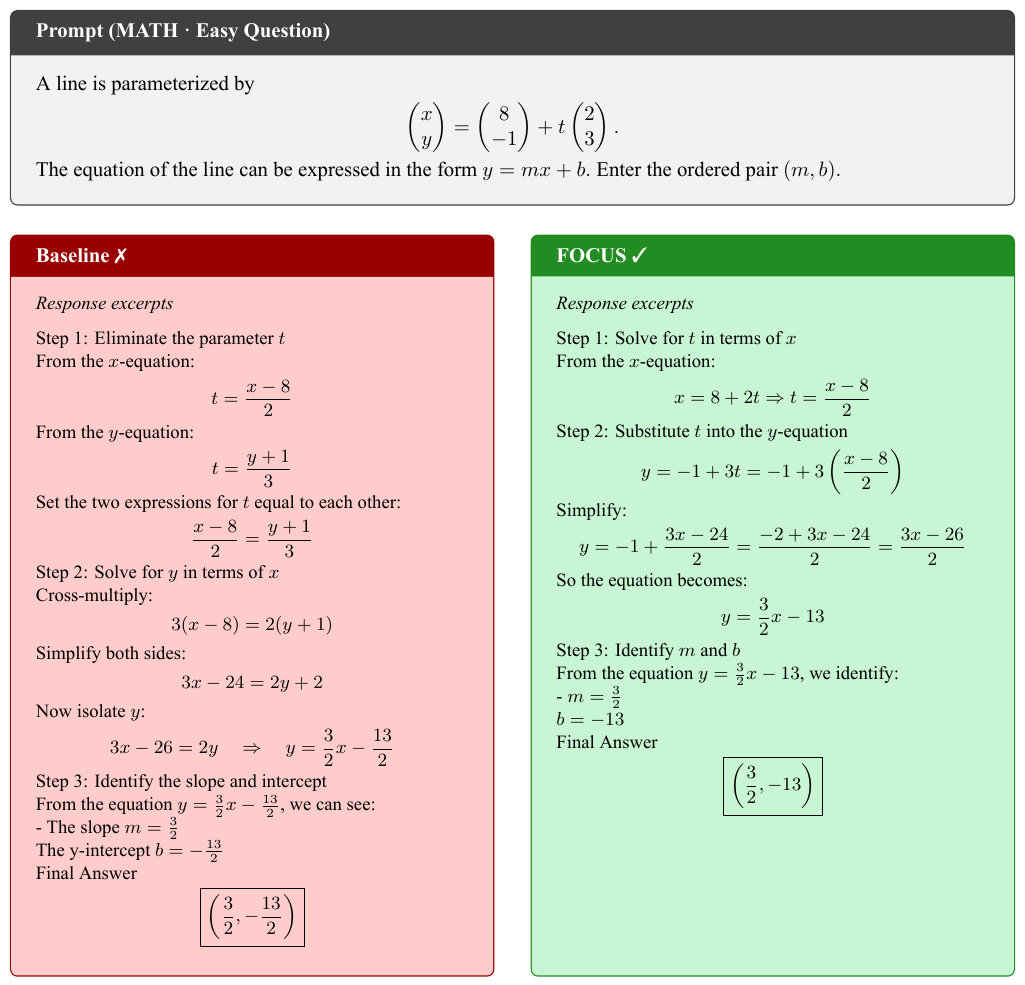}
  \caption{Baseline and FOCUS response excerpts on MATH (example 1).}
  \label{fig:downstream_case_1}
\end{figure}

\clearpage
\begin{figure}[!htbp]
  \centering
  \includegraphics[page=2,width=\linewidth,height=0.85\textheight,keepaspectratio,trim=24bp 284bp 24bp 24bp,clip]{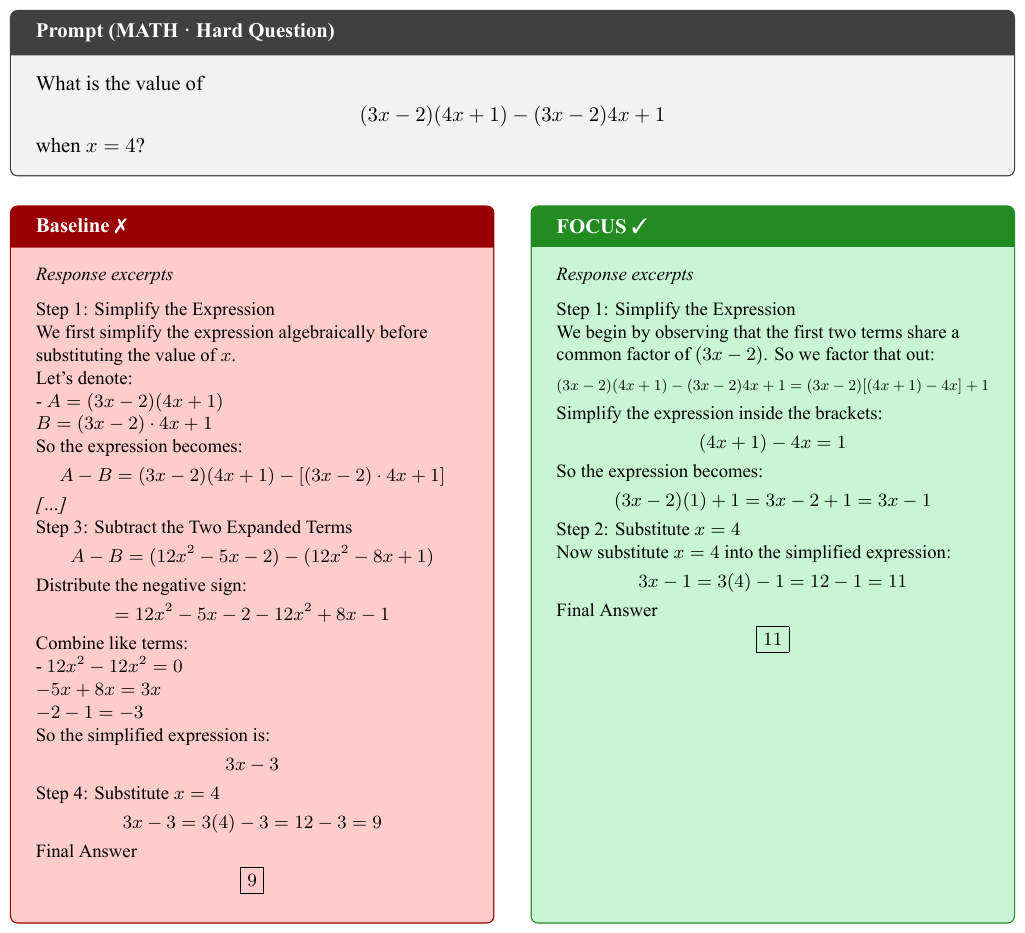}
  \caption{Baseline and FOCUS response excerpts on MATH (example 2).}
  \label{fig:downstream_case_2}
\end{figure}

\clearpage
\begin{figure}[!htbp]
  \centering
  \includegraphics[page=3,width=\linewidth,height=0.85\textheight,keepaspectratio,trim=24bp 198bp 24bp 24bp,clip]{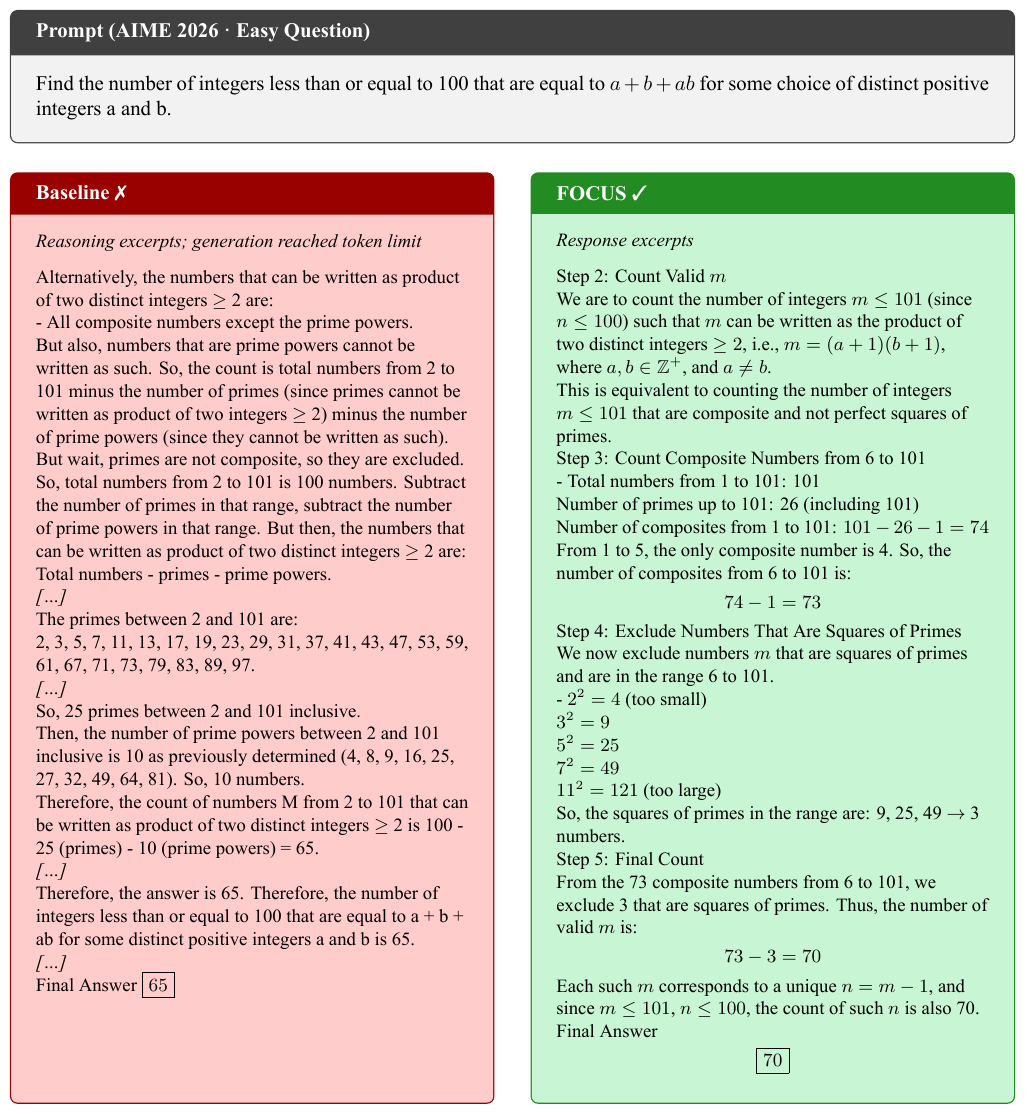}
  \caption{Baseline and FOCUS response excerpts on AIME 2026 (example 1).}
  \label{fig:downstream_case_3}
\end{figure}

\clearpage
\begin{figure}[!htbp]
  \centering
  \includegraphics[page=4,width=\linewidth,height=0.85\textheight,keepaspectratio,trim=24bp 211bp 24bp 24bp,clip]{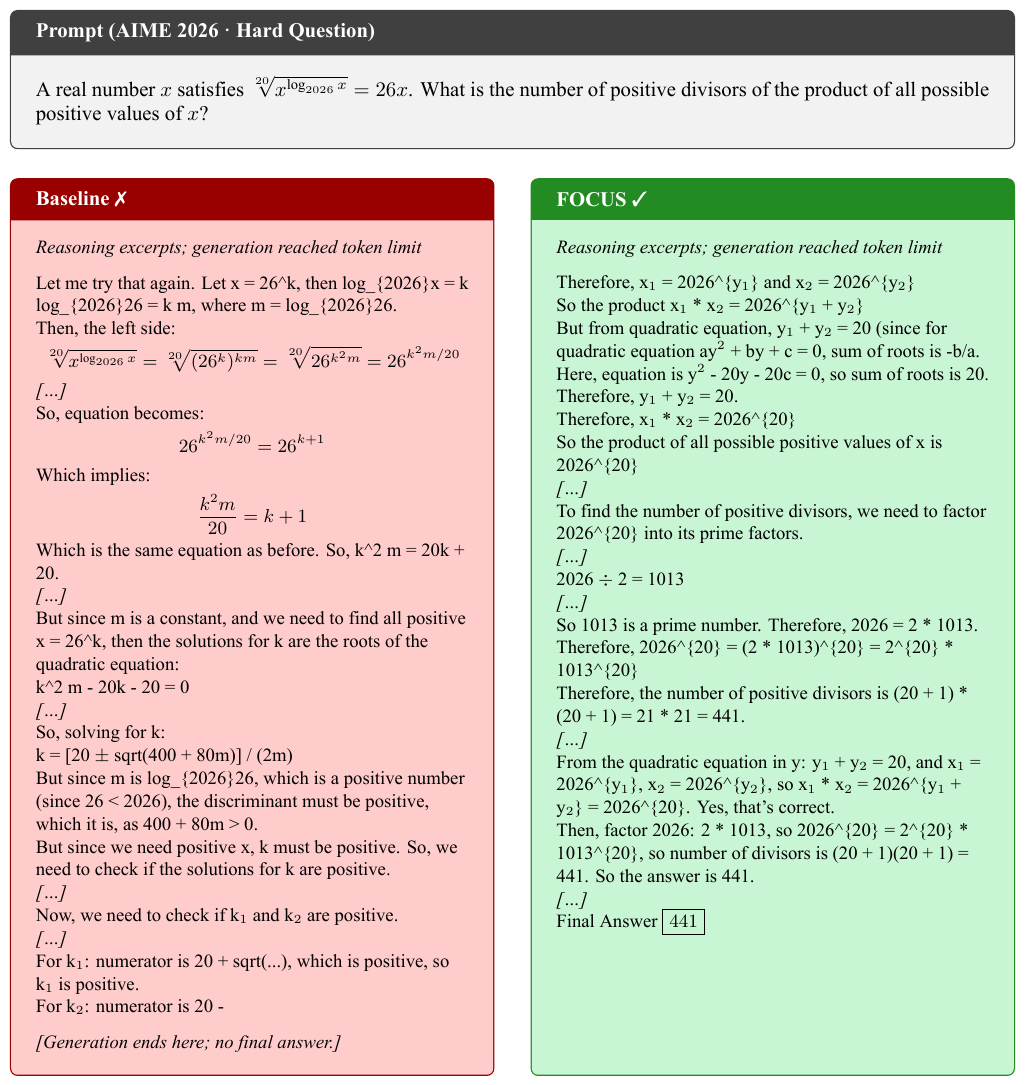}
  \caption{Baseline and FOCUS response excerpts on AIME 2026 (example 2).}
  \label{fig:downstream_case_4}
\end{figure}

\clearpage
\begin{figure}[!htbp]
  \centering
  \includegraphics[page=5,width=\linewidth,height=0.85\textheight,keepaspectratio,trim=24bp 246bp 24bp 24bp,clip]{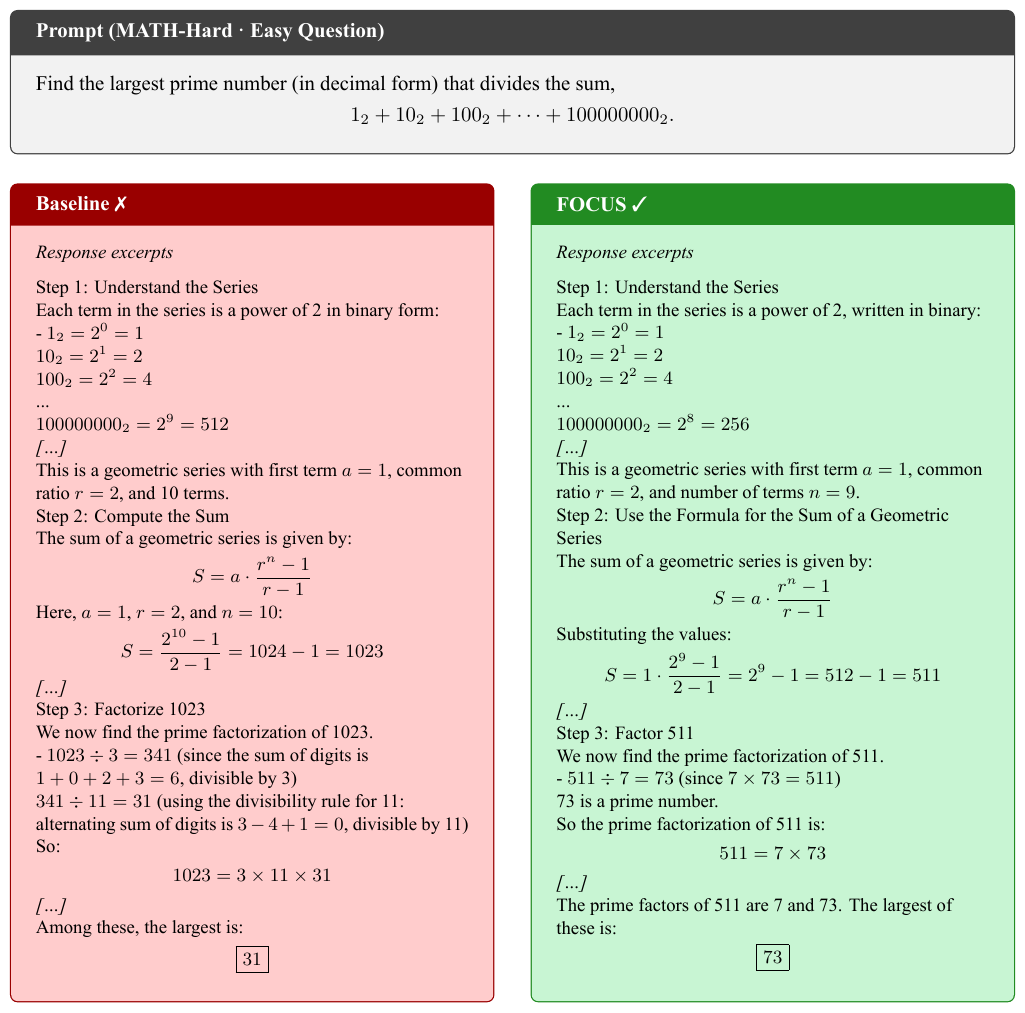}
  \caption{Baseline and FOCUS response excerpts on MATH-Hard (example 1).}
  \label{fig:downstream_case_5}
\end{figure}

\clearpage
\begin{figure}[!htbp]
  \centering
  \includegraphics[page=6,width=\linewidth,height=0.85\textheight,keepaspectratio,trim=24bp 170bp 24bp 24bp,clip]{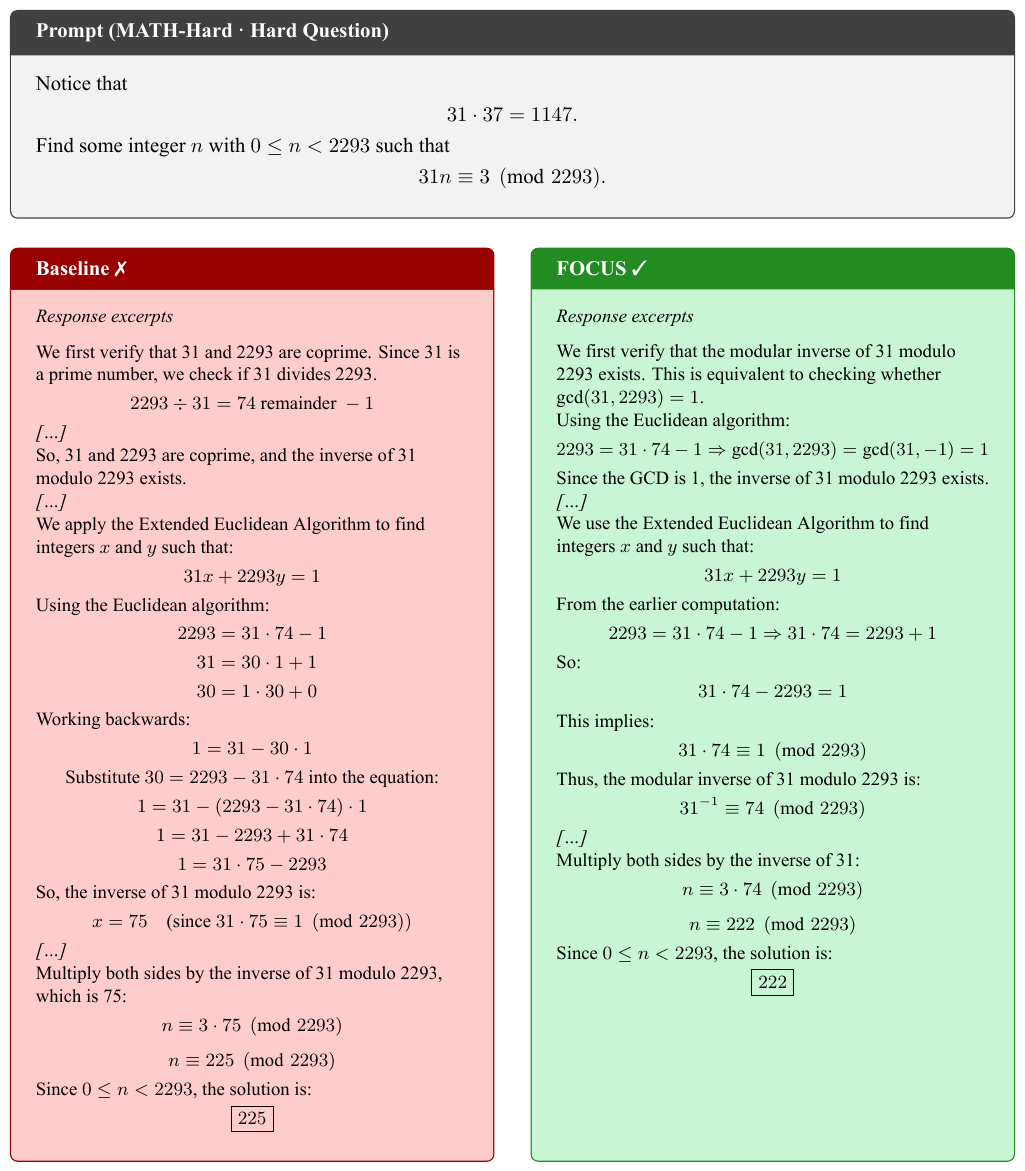}
  \caption{Baseline and FOCUS response excerpts on MATH-Hard (example 2).}
  \label{fig:downstream_case_6}
\end{figure}

\clearpage
\begin{figure}[!htbp]
  \centering
  \includegraphics[page=7,width=\linewidth,height=0.85\textheight,keepaspectratio,trim=24bp 183bp 24bp 24bp,clip]{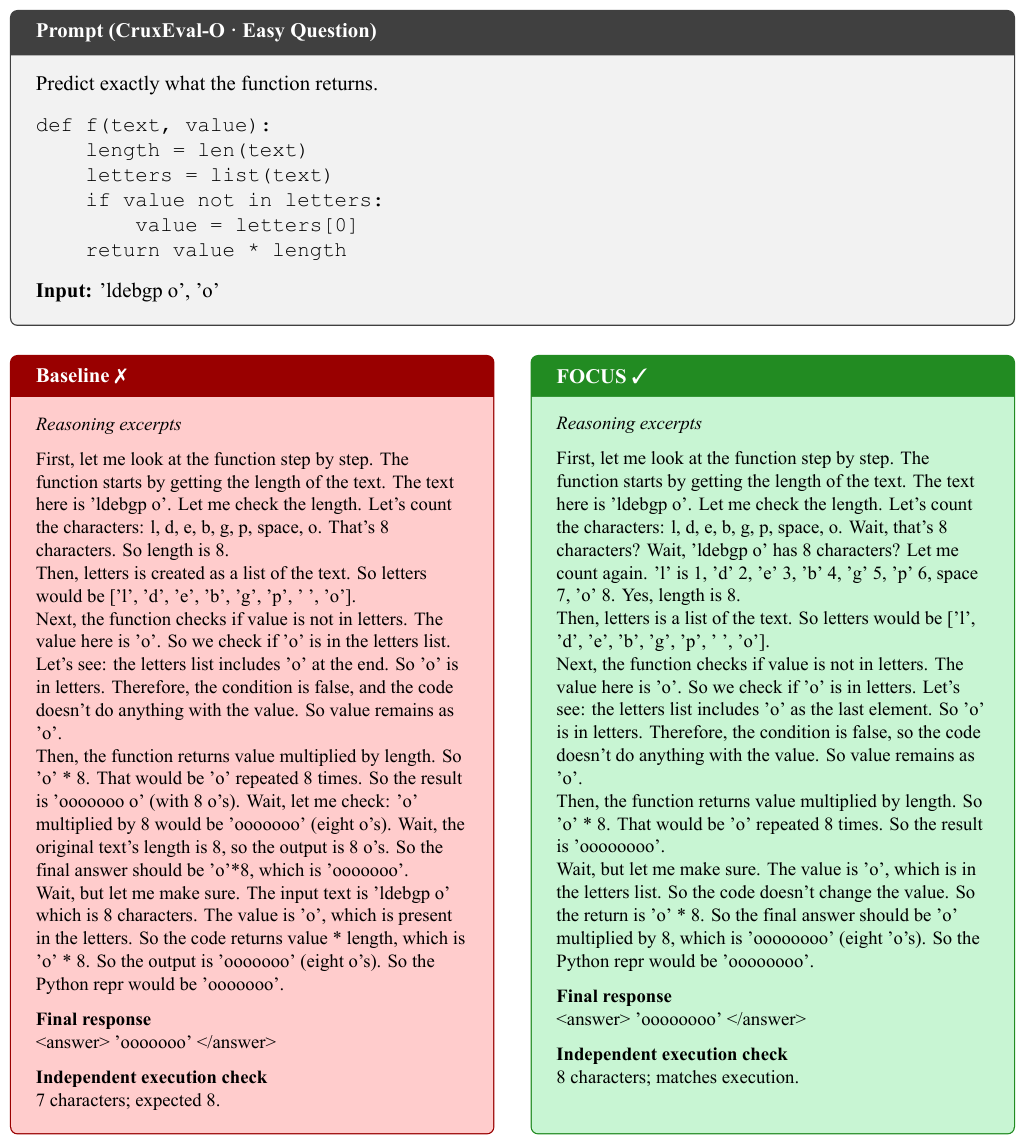}
  \caption{Baseline and FOCUS response excerpts on CruxEval-O (example 1).}
  \label{fig:downstream_case_7}
\end{figure}

\clearpage
\begin{figure}[!htbp]
  \centering
  \includegraphics[page=8,width=\linewidth,height=0.85\textheight,keepaspectratio,trim=24bp 183bp 24bp 24bp,clip]{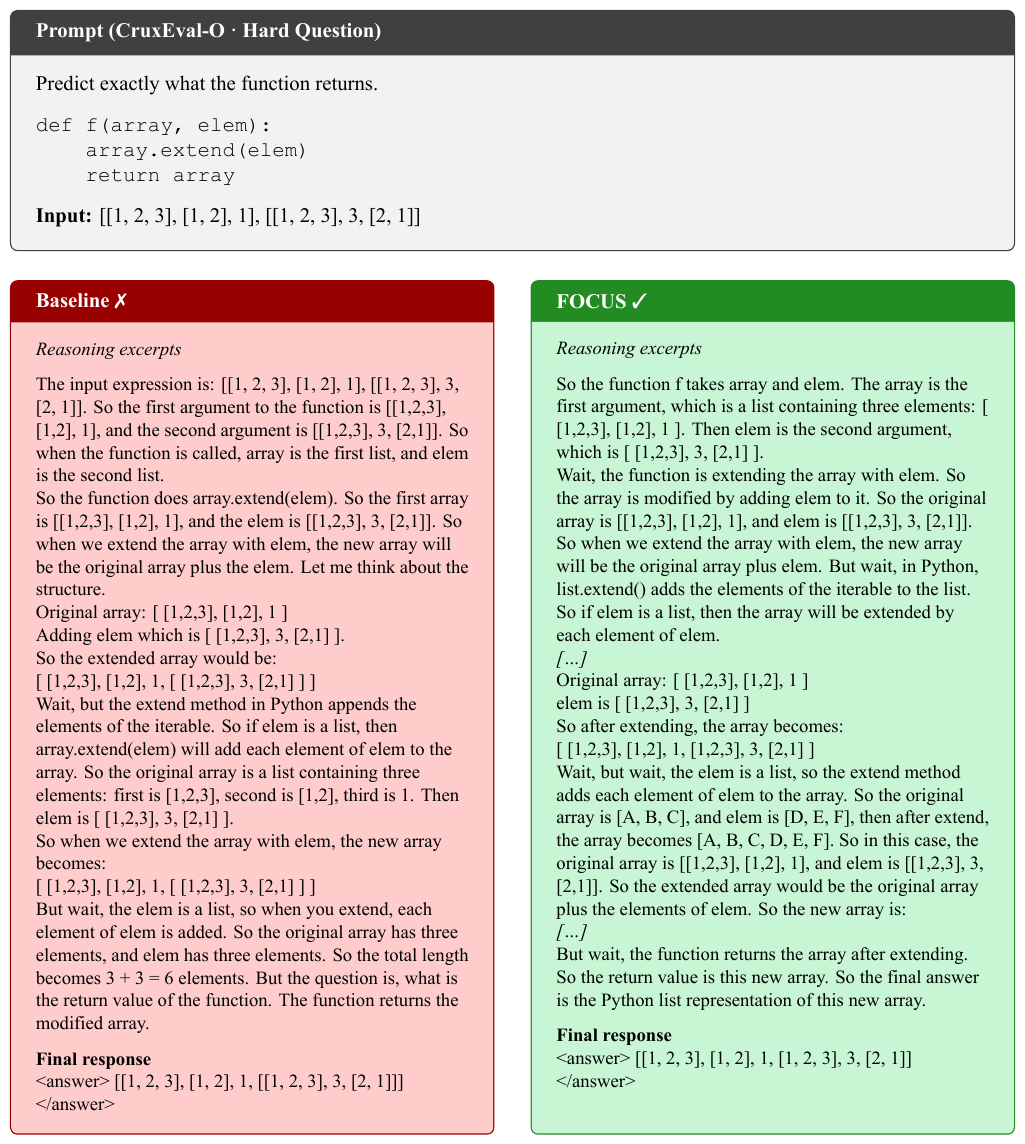}
  \caption{Baseline and FOCUS response excerpts on CruxEval-O (example 2).}
  \label{fig:downstream_case_8}
\end{figure}

\clearpage
\begin{figure}[!htbp]
  \centering
  \includegraphics[page=9,width=\linewidth,height=0.85\textheight,keepaspectratio,trim=24bp 138bp 24bp 24bp,clip]{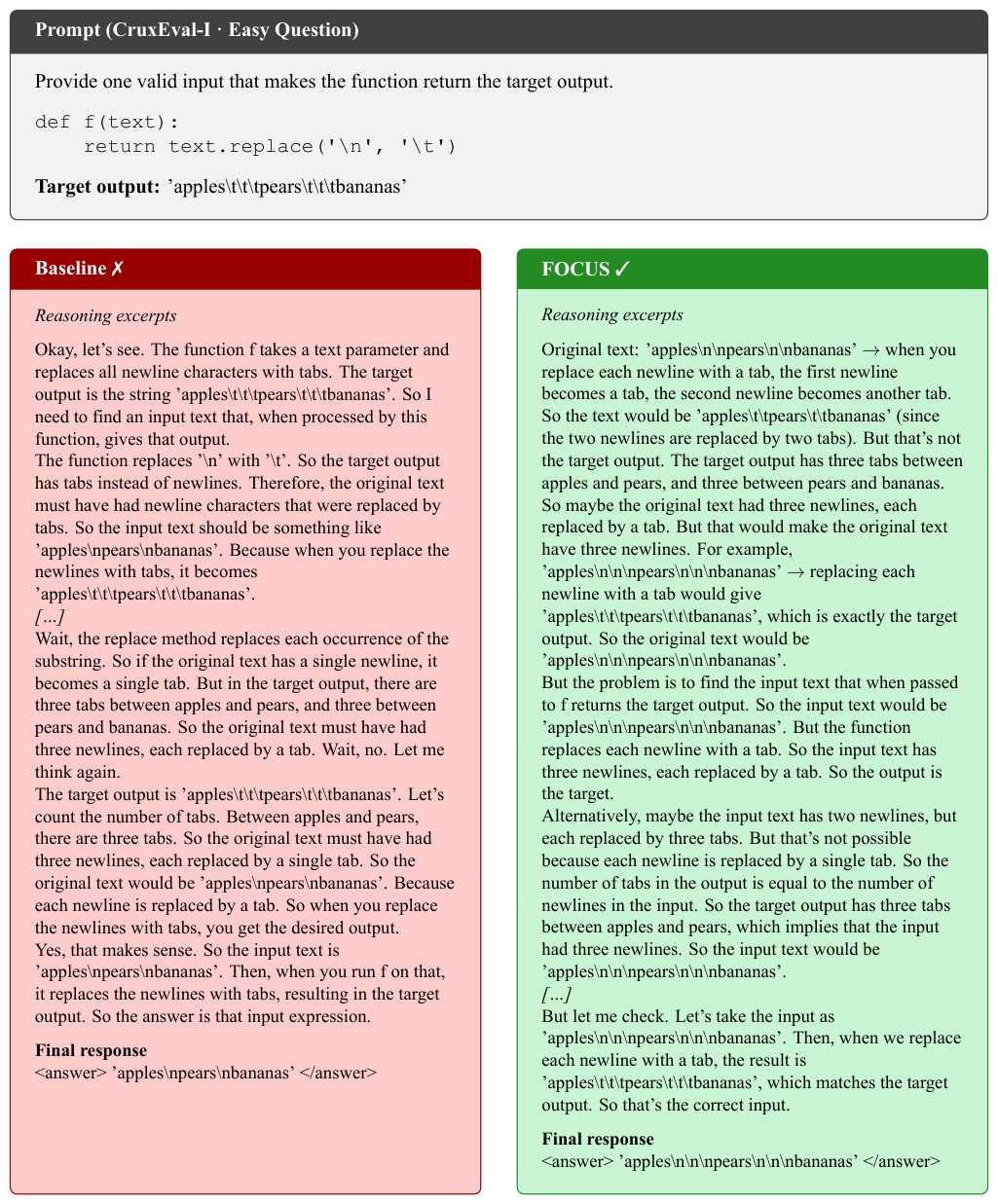}
  \caption{Baseline and FOCUS response excerpts on CruxEval-I (example 1).}
  \label{fig:downstream_case_9}
\end{figure}

\clearpage
\begin{figure}[!htbp]
  \centering
  \includegraphics[page=10,width=\linewidth,height=0.85\textheight,keepaspectratio,trim=24bp 27bp 24bp 24bp,clip]{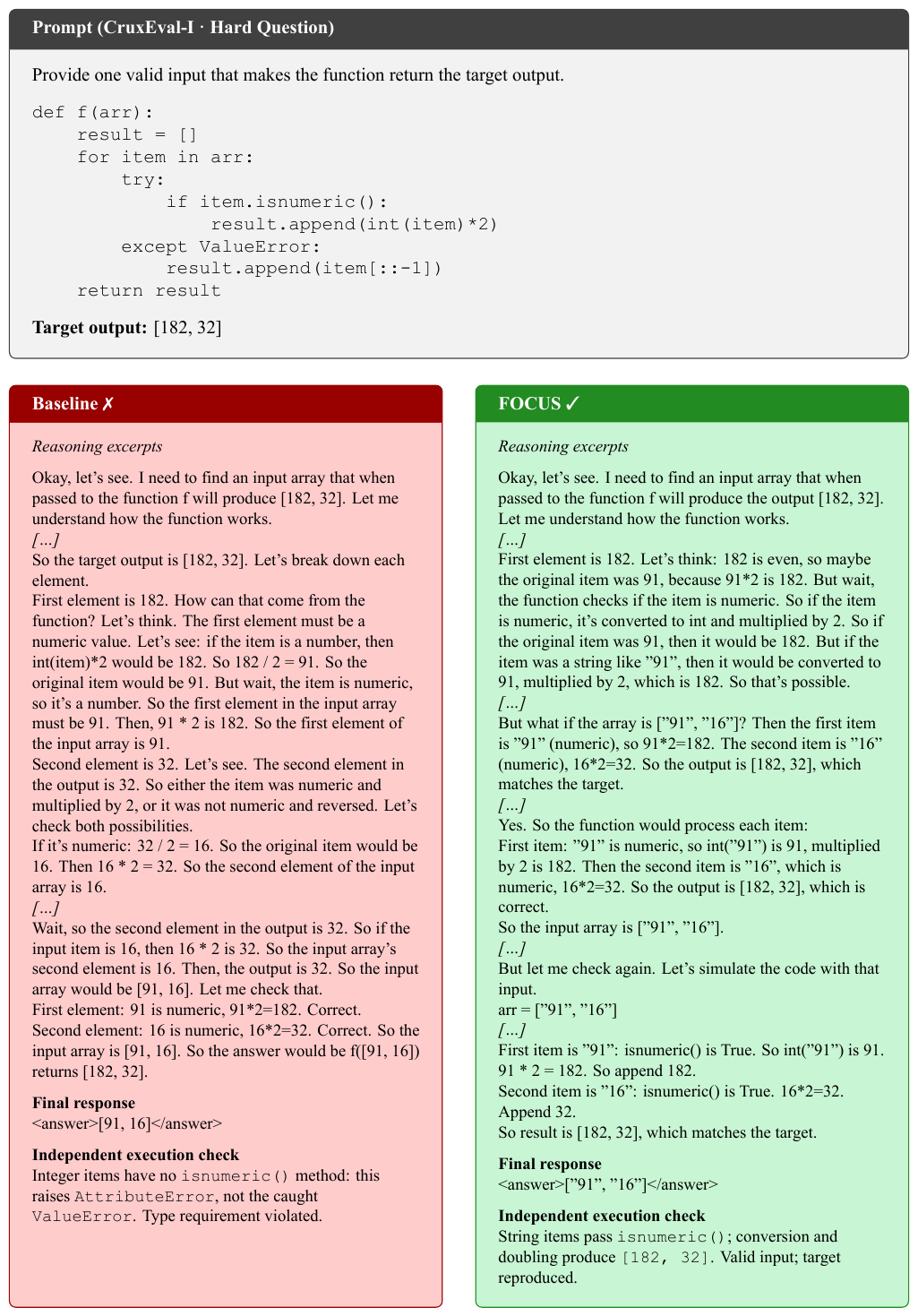}
  \caption{Baseline and FOCUS response excerpts on CruxEval-I (example 2).}
  \label{fig:downstream_case_10}
\end{figure}

\end{document}